%% file: 541.tex
\documentclass[a4paper,11pt,english]{article}

\pdfoutput=1

\usepackage{cite}
\usepackage{epsfig}
\usepackage{babel}
\usepackage{a4wide}
\usepackage{amsmath}
\usepackage{amsthm}
\usepackage{amssymb}
\usepackage{units}
\usepackage{booktabs}
\usepackage{setspace}
\usepackage[a4paper,left=15mm,right=10mm, top=3.5cm, bottom=2cm]{geometry}
\usepackage{ifthen}
\usepackage{color}
\usepackage{framed}
\usepackage{bm}
\usepackage{titlesec}
\usepackage{tocloft}
\makeatletter
\renewcommand{\@biblabel}[1]{\quad#1.}
\makeatother

\input{new_commands}

\newcommand{\Nd}{\textsc{Normal}}
\newcommand{\Pd}{\textsc{Poisson}}
\newcommand{\ddthetai}{\frac{\partial}{\partial \theta_i}}
\newcommand{\ddwi}{\frac{\partial}{\partial w_i}}
\newcommand{\ddbi}{\frac{\partial}{\partial b_i}}
\newcommand{\ddwij}{\frac{\partial}{\partial w_{ij}}}

\newcommand{\dd}[1]{\frac{\partial}{\partial #1}}

\newboolean{showToDos}

\setboolean{showToDos}{true}        

\newcommand{\bz}{\mathbf{z}}
\newcommand{\bx}{\mathbf{x}}
\newcommand{\bw}{\mathbf{w}}

\begin{document}
\title{
 \LARGE
Network Plasticity as Bayesian Inference
\\[5mm]}
\author{David Kappel$^{1,2}$, Stefan Habenschuss$^{1,2}$, Robert Legenstein${^1}$ , Wolfgang Maass$^1$
\\[3mm]
\\
\hspace{1pt}
$^1$Institute for Theoretical Computer Science, Graz University of Technology, 
A-8010 Graz, Austria
\\[2mm]
\hspace{1pt}
$^2$These authors contributed equally to this work.
\\[2mm]
\hspace{1pt}
Corresponding author: kappel@igi.tugraz.at
\hspace{1pt}
\\
\url{http://www.igi.tugraz.at/kappel}
}


\maketitle

\begin{abstract}
General results from statistical learning theory suggest to understand not only brain computations, but also brain plasticity as probabilistic inference. But a model for that has been missing. We propose that inherently stochastic features of synaptic plasticity and spine motility enable cortical networks of neurons to carry out probabilistic inference by sampling from a posterior distribution of network configurations. This model provides a viable alternative to existing models that propose convergence of parameters to maximum likelihood values. It explains how priors on weight distributions and connection probabilities can be merged optimally with learned experience, how cortical networks can generalize  learned information so well to novel experiences, and how they can compensate continuously for unforeseen disturbances of the network. The resulting new theory of network plasticity explains from a functional perspective a number of experimental data on stochastic aspects of synaptic plasticity that previously appeared to be quite puzzling.
\end{abstract}


\vspace{20mm}




\section{Introduction}

We reexamine in this article the conceptual and mathematical framework for understanding the organization of plasticity in networks of neurons in the brain. We will focus on synaptic plasticity and network rewiring (spine motility) in this article, but our framework is also applicable to other network plasticity processes. One commonly assumes, that plasticity moves network parameters $\ve \theta$ (such as synaptic connections between neurons and synaptic weights) to values $\ve \theta^*$ that are optimal for the current computational function of the network. In learning theory, this view is made precise for example as maximum likelihood learning, where model parameters $\ve \theta$ are moved to values $\ve \theta^*$ that maximize the fit of the resulting internal model to the inputs $\ve x$ that impinge on the network from its environment (by maximizing the likelihood of these inputs $\ve x$). The convergence to $\ve \theta^*$ is often assumed to be facilitated by some external regulation of learning rates, that reduces the learning rate when the network approaches an optimal solution.

This view of network plasticity has been challenged on several grounds. From the theoretical perspective it is problematic because in the absence of an intelligent external controller it is likely to lead to overfitting of the internal model to the inputs $\ve x$ it has received, thereby reducing its capability to generalize learned knowledge to new inputs. Furthermore, networks of neurons in the brain are apparently exposed to a multitude of internal and external changes and perturbations, to which they have to respond quickly in order to maintain stable functionality. In addition, experimental data suggest that these networks are simultaneously able to maintain structural constraints and rules such as the empirically found connection probability between specific types of neurons, and heavy-tailed distributions of synaptic weights \cite{SongETAL:05,BuzsakiMizuseki:14}, which are in conflict with maximum likelihood learning models. Other experimental data point to surprising ongoing fluctuations in dendritic spines and spine volumes, to some extent even in the adult brain \cite{HoltmaatSvoboda:09} and  in the absence of synaptic activity \cite{YasumatsuETAL:08}. Also a significant portion of axonal side branches and axonal boutons were found to appear and disapper within a week in adult visual cortex, even in the absence of imposed learning and lesions \cite{StettlerETAL:06}.
Furthermore surprising random drifts of tuning curves of neurons in motor cortex were observed \cite{RokniETAL:07}. 
Apart from such continuously ongoing changes in synaptic connections and tuning curves of neurons, massive changes in synaptic connectivity were found to accompany functional reorganization of primary visual cortex after lesions, see e.g. \cite{YamahachiETAL:09}. 

We therefore propose to view network plasticity as a process that continuously moves high-dimensional network parameters $\ve \theta$ within some low-dimensional manifold that represents a compromise between overriding structural rules and different ways of fitting the internal model to external inputs $\ve x$. We propose that ongoing stochastic fluctuations (not unlike Brownian motion) continuously drive network parameters $\ve \theta$ within such low-dimensional manifold. 
The primary conceptual innovation is the departure from the traditional view of learning as moving parameters to values $\ve \theta^*$ that represent optimal (or locally optimal) fits to network inputs $\ve x$. 
We show that our alternative view can be turned into a precise learning model within the framework of probability theory. This new model satisfies theoretical requirements for handling priors such as structural constraints and rules in a principled manner, that have previously already been formulated and explored in the context of artificial neural networks \cite{MacKay:92a,Bishop:06}, as well as more recent challenges that arise from probabilistic brain models \cite{PougetETAL:13}. The low-dimensional manifold of parameters $\ve \theta$ that becomes the new learning goal in our model can be characterized mathematically as the high probability regions of the posterior distribution $p^*(\ve \theta | \ve x)$ of network parameters $\ve \theta$. This posterior arises as product of a general prior $p_\mathcal{S} (\ve \theta)$ for network parameters (that enforces structural rules) with a term that describes the quality of the current internal model (e.g. in a predictive coding or generative modeling framework: the likelihood $p_{\mathcal{N}}(\ve x | \ve \theta)$ of inputs $\ve x$ for the current parameter values $\ve \theta$ of the network $\mathcal{N}$). More precisely, we propose that brain plasticity mechanisms are designed to enable brain networks to sample from this posterior distribution  $p^*( \ve \theta | \ve x)$ through inherent stochastic features of their molecular implementation.
In this way synaptic and other plasticity processes are able to carry out probabilistic (or Bayesian) inference through sampling from a posterior distribution that takes into account both structural rules and fitting to external inputs.
Hence this model provides a solution to the challenge of \cite{PougetETAL:13} to understand how posterior distributions of weights can be represented and learned by networks of neurons in the brain. 

This new model proposes to reexamine rules for synaptic plasticity. Rather than viewing trial-to-trial variability and ongoing fluctuations of synaptic parameters as the result of a suboptimal implementation of an inherently deterministic plasticity process, it proposes to model experimental data on synaptic plasticity by rules that consist of three terms: the standard (typically deterministic) activity-dependent (e.g., Hebbian or STDP) term that fits the model to external inputs, a second term that enforces structural rules (priors), and a third term that provides the stochastic driving force. This stochastic force enables network parameters to sample from the posterior, i.e., to fluctuate between different possible solutions of the learning task. The stochastic third term can be modeled by a standard formalism (stochastic Wiener process) that had been developed to model Brownian motion.
The first two terms can be modeled as drift terms in a stochastic process. A key insight is that one can easily relate details of the resulting more complex rules for the dynamics of network parameters $\ve \theta$, which now become stochastic differential equations, to specific features of the resulting posterior distribution $p^*(\ve \theta | \bx)$ of parameter vectors $\ve \theta$ from which the network samples. Thereby, this theory provides a new framework for relating experimentally observed details of local plasticity mechanisms (including their typically stochastic implementation on the molecular scale) to functional consequences of network learning. For example, one gets a theoretically founded framework for relating experimental data on spine motility to experimentally observed network properties, such as sparse connectivity, specific distributions of synaptic weights, and the capability to compensate against perturbations \cite{Marder:11}.

We demonstrate the resulting new style of modeling network plasticity in three examples. These examples demonstrate how previously mentioned functional demands on network plasticity, such as incorporation of structural rules, automatic avoidance of overfitting, and inherent and immediate compensation for network perturbances, can be accomplished through stochastic local plasticity processes. We focus here on 
common models for unsupervised learning in networks of neurons: generative models. We first develop the general learning theory for this class of models, and then describe applications to common non-spiking and spiking generative network models. 
Both
structural plasticity (see 
\cite{May:11,CaroniETAL:12} 
for reviews) and synaptic plasticity (STDP) are integrated into the resulting theory of network plasticity.

\section{Results}

We present a new theoretical framework for analyzing and understanding local plasticity mechanisms of networks of neurons in the brain as stochastic processes, that generate specific distributions $p(\ve \theta)$ of network parameters $\ve \theta$ over which these parameters fluctuate. This framework can be used to analyze and model many types of learning processes. We illustrate it here for the case of unsupervised learning, i.e., learning without a teacher or rewards. Obviously many learning processes in biological organisms are of this nature, especially learning processes in early sensory areas, and in other brain areas that have to provide and maintain on their own an adequate level of functionality, even in the face of internal or external perturbations. 

A common framework for modeling unsupervised learning in networks of neurons are generative models, which date 
back to the 19'th century, when Helmholtz proposed that perception could be understood as unconscious inference \cite{Hatfield:02}. Since then the hypothesis of the ``generative brain'' has been receiving considerable attention, fueling
interest in various aspects of the relation between Bayesian inference and the brain \cite{RaoETAL:02,DoyaETAL:07,PougetETAL:13}. 
The basic assumption of the ``Bayesian brain'' theory is that the
activity $\ve z$ of neuronal networks in the brain 
can be viewed as an internal model for 
hidden variables in the outside world that give rise to sensory experiences
$\ve x$ (such as the response $\ve x$ of auditory sensory neurons to spoken
words that are guessed by an internal model $\ve z$). The internal model $\ve z$ is usually assumed to be represented by the activity of neurons in the network, e.g., in terms of the firing rates of neurons, or in terms of
spatio-temporal spike patterns. A network $\mathcal{N}$ of stochastically firing neuron is modeled in this framework
by a probability distribution $p_{\mathcal{N}}(\bx, \bz | \ve \theta)$ that
describes the probabilistic relationships between $N$ input patterns $ \bx = \ve x^1, \ldots, \ve x^N $ and corresponding network responses $\bz = \ve z^1, \ldots, \ve z^N $, where $\ve \theta$ denotes the vector of network parameters that shape this distribution, e.g., via synaptic efficacies and network connectivity.
The marginal probability $p_{\mathcal{N}}(\bx | \ve\theta) = \sum_{\bz} p_{\mathcal{N}} (\bx, \bz  | \ve \theta)$ of the
actually occurring inputs $\bx = \ve x^1, \ldots, \ve x^N$ under the resulting internal model of the neural network $\mathcal{N}$ with parameters $\ve \theta$ can then be viewed as a measure for the agreement between this internal model (which carries out ``predictive coding'' \cite{WinklerETAL:12}) and its environment (which generates the inputs $\bx$). 

The goal of network learning is usually described in this probabilistic generative framework as finding parameter values $\ve \theta^*$ that maximize 
this agreement, or equivalently the likelihood of the inputs $\bx$ (maximum likelihood learning):
\begin{equation}\label{eq:ml_theta}
   \ve \theta^* = \arg \max_{\ve \theta} p_{\mathcal{N}}(\bx | \ve\theta).
\end{equation}

\noindent Locally optimal parameter solutions are usually determined by gradient ascent
on the data likelihood $p_{\mathcal{N}}(\bx | \ve \theta)$. 

\subsection{Learning a posterior distribution through stochastic synaptic plasticity}
\label{sec:learning_post}

In contrast, we assume here that not only a neural network $\mathcal{N}$, but also a prior $p_\mathcal{S}(\ve \theta)$ for its parameters are given. This prior $p_\mathcal{S}$ can encode both structural constraints (such as sparse connectivity) and structural rules (e.g., a heavy-tailed distribution of synaptic weights). Then the goal of network learning becomes: 
\begin{align}
\begin{split}
\parbox{0.8\textwidth}{
learn the posterior distribution $p^*(\ve \theta | \bx)$ defined (up to normalization) by $p_\mathcal{S} (\ve \theta) \cdot p_{\mathcal{N}} (\bx | \ve \theta)$\;.}
\end{split}
\label{equ:produces_parameter}
\end{align}

\noindent The patterns $\bx = \ve x^1, \ldots, \ve x^N$ are assumed here to be regularly reoccurring network inputs. 

\begin{figure}
\centering
\includegraphics{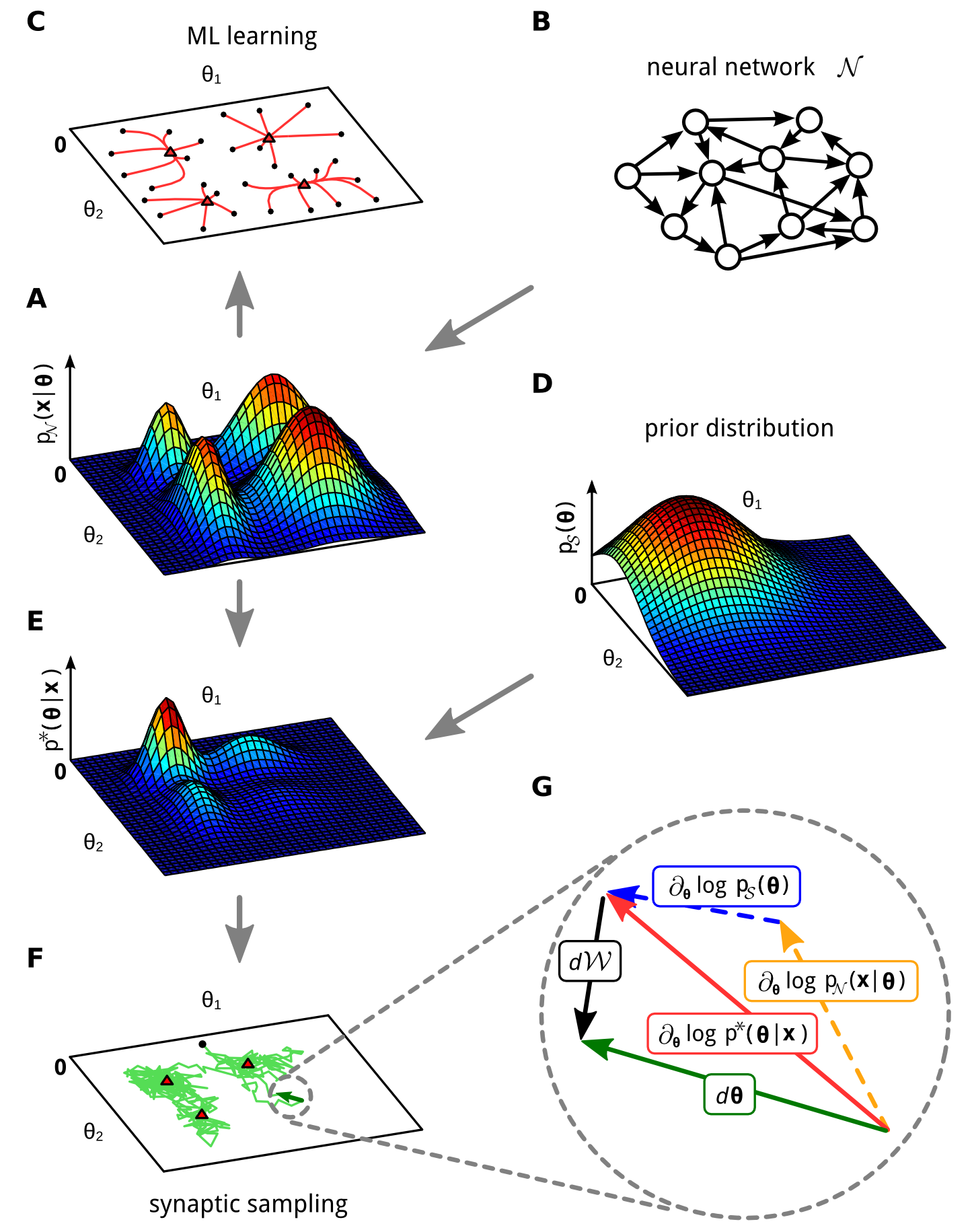}
\caption{{\bf Maximum likelihood (ML) learning vs.~synaptic sampling.}
(A,B,C) Illustration of ML learning for two parameters $\ve\theta=\{\theta_1,\theta_2\}$ of a neural network $\mathcal{N}$. 
(A) 3D plot of an example likelihood function.  For a fixed set of inputs $\bx$ it assigns a probability density (amplitude on z-axis) to each parameter setting $\ve \theta$. (B) This likelihood function is defined by some
underlying neural network $\mathcal{N}$. (C) Multiple trajectories along the gradient of the likelihood function in (A). The parameters are initialized at random initial values (black dots) and then follow the gradient 
to a local maximum (red triangles). (D) 
Example for a prior that prefers small values for $\ve\theta$. (E) The posterior that results as product of the prior (D) and the likelihood (A). (F) A single trajectory of synaptic sampling from the posterior (E), starting at the black dot. The parameter vector $\ve \theta$ fluctuates between different solutions, the visited values cluster near local optima (red triangles).
(G) Cartoon illustrating 
the dynamic forces (plasticity rule \eqref{eqn:ml-mcmc2}) that enable the network to sample from the posterior distribution $p^*(\ve \theta | \bx)$ in (E).
Magnification of 
one synaptic sampling step $d \ve \theta$ of the trajectory in (F) (green). The three forces acting on $\ve \theta$: the deterministic drift term (red) is
directed to the next local maximum (red triangle), it consists
of the first two terms in Eq.~\eqref{eqn:ml-mcmc2}; the stochastic diffusion term $d \wiener$ (black) has a random direction. See \nameref{sec:supplement} for figure details.
}
\label{fig:model-illustration}
\end{figure}

A key insight (see Fig.~\ref{fig:model-illustration} for an illustration) is that stochastic local plasticity rules for the parameters $\theta_i$ enable a network to achieve the learning goal \eqref{equ:produces_parameter}: The distribution of network parameters $\ve \theta$ will converge after a while to the posterior distribution $p^*(\ve \theta) = p^* (\ve \theta | \bx)$ -- and produce samples from it --
if each network parameter $\theta_i$ obeys the dynamics
\begin{align}
 d \theta_i \;=\; b \left( \ddthetai \,\log p_\mathcal{S}(\ve \theta) + \ddthetai \log p_{\mathcal{N}} (\bx | \ve \theta) \right)\, dt  \;+ \; \sqrt{2 b} \, d \wiener_i \;,
 \label{eqn:ml-mcmc2}
\end{align}
where the learning rate $b>0$ controls the speed of the parameter dynamics. 
Eq.~\eqref{eqn:ml-mcmc2} is a stochastic differential equation (see \cite{Gardiner:04}),
which differs from commonly considered differential equations through the stochastic term $d \wiener_i$ that describes
infinitesimal stochastic increments and decrements of a Wiener process $\wiener_i$.
A Wiener process is a standard model for Brownian motion in one dimension (more precisely: the limit of a random walk with infinitesimal step size and normally distributed increments $\wiener_i^t - \wiener_i^s \sim \Nd(0, t - s)$ between times $t$ and $s$).
Thus in an approximation of \eqref{eqn:ml-mcmc2} for discrete time steps $\Delta t$ the term $d \wiener_i$ can be replaced by Gaussian noise with variance $\Delta t$ (see Eq.~\eqref{eq:onlinediscrete_mt}). Note that Eq.~\eqref{eqn:ml-mcmc2} does not have a single solution $\theta_i(t)$, but a continuum of stochastic sample paths (see Fig.~\ref{fig:model-illustration}F for an example) that each describe one possible time course of the parameter $\theta_i$. 

 Rigorous mathematical results based on Fokker-Planck equations (see \nameref{sec:methods} and \nameref{sec:supplement} for details) allow us to infer from the stochastic local dynamics of the parameters $ \theta_i$ given by a stochastic differential equation of the form \eqref{eqn:ml-mcmc2} the probability that the parameter vector $\ve \theta$ can be found after a while in a particular region of the high-dimensional space in which it moves. The key result is that for the case of the stochastic dynamics according to Eq.~\eqref{eqn:ml-mcmc2} this probability is equal to the posterior $p^*(\ve \theta | \bx)$ given by Eq.~\eqref{equ:produces_parameter}. 
Hence the stochastic dynamics \eqref{eqn:ml-mcmc2} of network parameters $\theta_i$ enables a network to achieve the learning goal \eqref{equ:produces_parameter}: to learn the posterior distribution $p^*(\ve \theta | \bx)$. This posterior distribution is not represented in the network through any explicit neural code, but through its stochastic dynamics, as the unique stationary distribution of a Markov process from which it samples continuously. In particular,  
if most of the mass of this posterior distribution is concentrated on some low-dimensional manifold, the network parameters $\ve \theta$ will move most of the time within this low-dimensional manifold. 
Since this realization of the posterior distribution $p^*(\ve \theta | \bx)$ is achieved by sampling from it, we refer to this model \eqref{eqn:ml-mcmc2} (in the case where the parameters $\theta_i$ represent synaptic parameters) as \emph{synaptic sampling}.

The stochastic term $ d \wiener_i $ in Eq.~\eqref{eqn:ml-mcmc2} 
provides a simple integrative model for
a multitude of biological and biochemical stochastic processes that effect the efficacy of a synaptic connection. The mammalian postsynaptic density comprises over $1000$ different types of proteins \cite{CobaETAL:09}. Many of those proteins that effect the amplitude of postsynaptic potentials and synaptic plasticity, for example NMDA receptors, occur in small numbers, and are subject to Brownian motion within the membrane \cite{RibraultETAL:11}. 
In addition, the turnover of important scaffolding proteins in the postsynaptic density such as PSD-95, which clusters glutamate receptors and is thought to have a substantial impact on synaptic efficacy, is relatively fast, on the time-scale of hours to days, depending on developmental state and environmental condition \cite{GrayETAL:06}. Also the volume of spines at dendrites, which is assumed to be directly related to synaptic efficacy \cite{EngertBonhoeffer:99,HoETAL:11} is reported to fluctuate continuously, even in the absence of synaptic activity \cite{YasumatsuETAL:08}. 
Furthermore the stochastically varying internal states of multiple interacting biochemical signaling pathways in the postsynaptic neuron are likely to effect synaptic transmission and plasticity \cite{BhallaIyengar:99}. 

The contribution of the stochastic term  $ d \wiener_i $ in \eqref{eqn:ml-mcmc2} can be scaled by a temperature parameter $\sqrt{T}$, where $T$ can be any positive number. The resulting stationary distribution of $\ve \theta$ is proportional to $p^*(\ve \theta)^{\frac{1}{T}}$, so that the dynamics of the stochastic process can be described by the energy landscape $\frac{\log p^*(\ve \theta)}{T}$. For high values of $T$ this energy landscape is flattened, i.e., the main modes of $p^*(\ve \theta)$ become less pronounced. For $T \rightarrow 0$ the dynamics of $\ve \theta$ approaches a deterministic process and converges to the next local maximum of $p^*(\ve \theta)$. Thus the learning process approximates for low values of $T$ maximum a posteriori (MAP) inference \cite{Bishop:06}.
We propose that this temperature parameter $T$ is regulated in biological networks of neurons in dependence of the developmental state, environment, and behavior of an organism. One can also accommodate a modulation of the dynamics of each individual parameter $\theta_i$ by a learning rate $b(\theta_i)$ that depends on its current value (see \nameref{sec:methods}). 
\vspace{0.5cm}

\subsection{Online synaptic sampling}

For online learning one assumes that the likelihood $p_{\mathcal{N}} (\bx | \ve \theta) =  p_{\mathcal{N}} (\ve x^1, \ldots,\ve x^N | \ve \theta)$ of the network inputs can be factorized:
\begin{equation}\label{eq:factorized_joint}
p_{\mathcal{N}} (\ve x^1, \ldots,\ve x^N | \ve \theta) = \prod^N_{n=1} p_{\mathcal{N}} (\ve x^n | \ve \theta), 
\end{equation}
i.e., each network input $\ve x^n$ can be explained as being drawn individually from $p_{\mathcal{N}} (\ve x^n | \ve \theta)$, independently from other inputs.  

The weight update rule \eqref{eqn:ml-mcmc2} depends on all inputs $\bx = \ve x^1, \ldots, \ve x^N$, hence synapses have to keep track of the whole set of all network inputs for the exact dynamics (batch learning). In an online scenario, we assume that only the current network input $\ve x^n$ is available for synaptic sampling. One then arrives at the following online-approximation to \eqref{eqn:ml-mcmc2}
\begin{align}
 d \theta_i \;=\; b \left(\ddthetai \log p_\mathcal{S}(\ve \theta) + N \,\frac{\partial}{\partial \theta_i} \,\log p_{\mathcal{N}}(\ve x^n | \ve \theta) \right)\,  dt  \;+ \; \sqrt{2 b\,} \, d \wiener_i \;.
 \label{eqn:ml-mcmc_approx}
\end{align}
Note the additional factor $N$ in
the rule. 
It compensates for the $N$-fold summation of the first and last term in \eqref{eqn:ml-mcmc_approx} when one moves through all $N$ inputs $\ve x^n$.
Although convergence to the correct posterior distribution cannot be guaranteed theoretically for
this online rule, we show in \nameref{sec:methods} that the rule is a
reasonable approximation to the batch-rule \eqref{eqn:ml-mcmc2}.
Furthermore, all subsequent simulations are based on this online rule,
which demonstrates the viability of this approximation.

\subsection{Relationship to maximum likelihood learning}

Typically, synaptic plasticity in generative network models is modeled
as maximum likelihood learning. Time is often discretized into small
discrete time steps $\Delta t$. For gradient-based approaches the parameter change $\Delta \theta_i^{ML}$ is then given by the gradient of the log likelihood multiplied with some learning rate $\eta$:
\begin{equation}\label{eq:delta}
  \Delta \theta_i^{ML} = \eta \ddthetai \log p_{\mathcal{N}}(\ve x^n |\ve \theta) \;.
\end{equation}
To compare this maximum likelihood update with synaptic sampling, we consider a version of
the parameter dynamics \eqref{eqn:ml-mcmc_approx} for discrete time (see \nameref{sec:methods} for a derivation):
\begin{equation}\label{eq:onlinediscrete_mt}
 \Delta \theta_i = \eta \left(\ddthetai \log p_\mathcal{S}(\ve \theta) + N\,\ddthetai \log p_{\mathcal{N}}(\ve x^n |\ve \theta)\right) + \sqrt{2 \eta } \, \nu_i^t\;\;,
\end{equation}
where the learning rate $\eta$ is given by $\eta = b \, \Delta t$ and $\nu_i^t$ denotes Gaussian noise with zero mean and variance
$1$, drawn independently for each parameter $\theta_i$ and each update time $t$. 
We see that the maximum likelihood update \eqref{eq:delta} becomes one term
in this online version of synaptic sampling. Equation \eqref{eq:onlinediscrete_mt} is a special case of the online Langevin sampler that was introduced in \cite{WellingTeh:11}.

The first term $\ddthetai \, \log p_\mathcal{S}(\ve \theta)$ in \eqref{eq:onlinediscrete_mt} arises from the prior $p_\mathcal{S}(\ve \theta)$, and has apparently not been considered in previous rules for synaptic plasticity. An additional novel component is the Gaussian noise term $ \nu_i^t $ (see also Fig.~\ref{fig:model-illustration}G). It arises because the accumulated impact of the Wiener process $\wiener_i$ over a time interval of length $\Delta t$ is distributed according to a normal distribution with variance $\Delta t$. In contrast to traditional maximum likelihood optimization based on additive noise for escaping local optima, this noise term is not scaled down when learning approaches a local optimum. This ongoing noise is essential for enabling the network to sample from the posterior distribution $p^*(\ve \theta)$ via continuously ongoing synaptic plasticity (see Fig.~\ref{fig:model-illustration}F and Fig.~\ref{fig:self-repair}C).

\subsection{Synaptic sampling improves the generalization capability of a neural network}
\label{sec:generalization}

\begin{figure}
\centering
\includegraphics{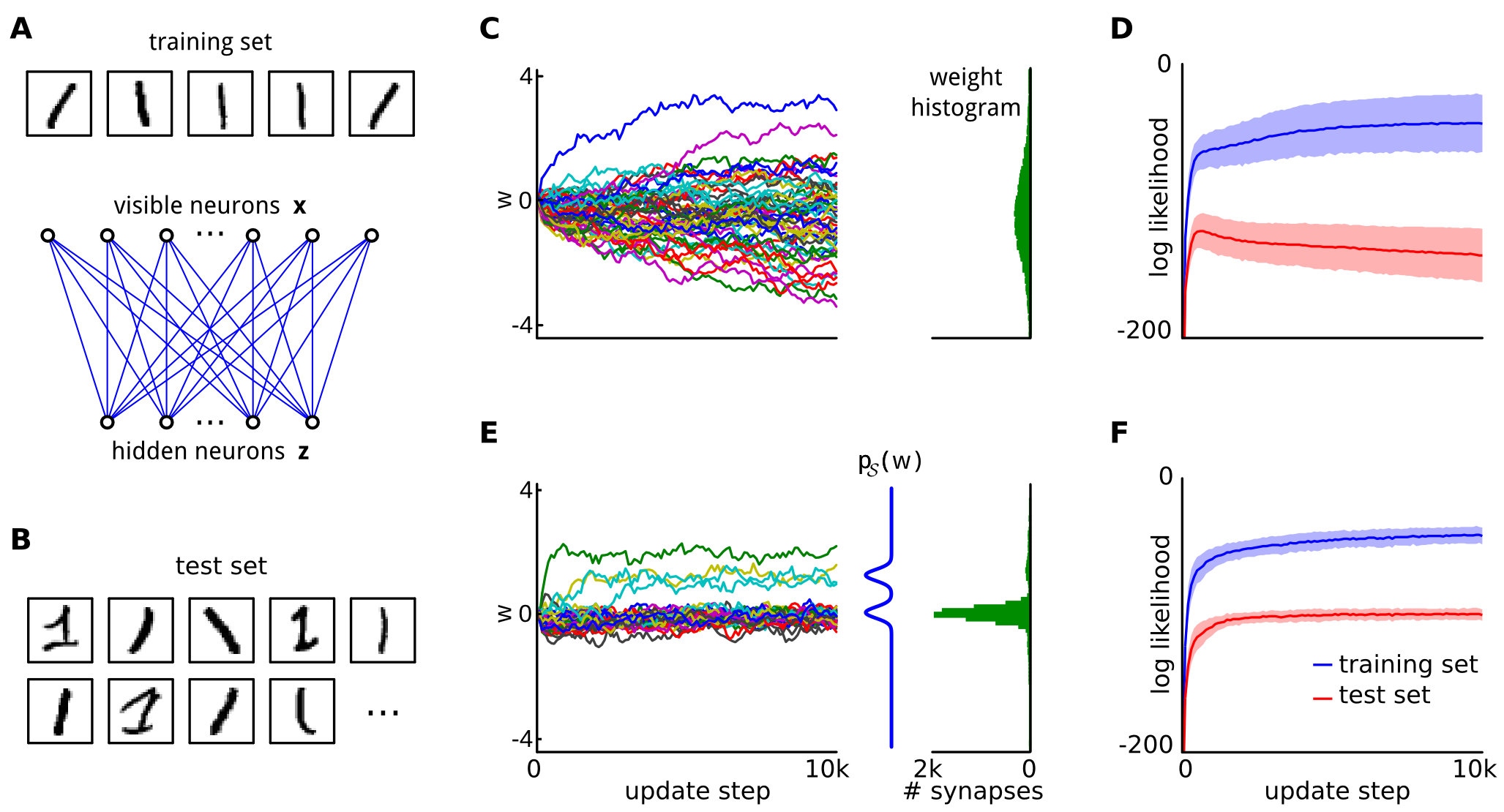}
\caption{{\bf Priors for synaptic weights improve generalization capability.}
(A) The training set, consisting of five samples of a handwritten \textit{1}. 
Below a cartoon illustrating the network architecture of the restricted Boltzmann machine (RBM), composed of a layer of 784 visible neurons $\bx$ and a layer of 9 hidden neurons $\bz$. (B) Examples from the test set. It contains many different styles of writing that are not part of the training set.  (C) Evolution of 50 randomly selected synaptic weights throughout learning (on the training set). 
The weight histogram (right) shows the distribution of synaptic weights at the end of learning. 80 histogram bins were equally spaced between -4 and 4.
(D) Performance of the network in terms of log likelihood on the training set (blue) and on the test set (red) throughout learning. Mean values over 100 trial runs are shown, shaded area indicates std. The performance on the test set initially increases but degrades for prolonged learning. (E) Evolution of 50 weights for the same network but with a bimodal prior. The prior $p_\mathcal{S}(w)$ is indicated by the blue curve. Most synaptic weights settle in 
the mode around $0$, but a few larger weights also emerge and stabilize in the larger mode.
Weight histogram (green) as in (C). 
(F) The log likelihood on the test set 
maintains 
a constant high value throughout the whole learning session, compare to (D).
}
\label{fig:overfitting}
\end{figure}

The previously described theory for learning a posterior distribution over parameters $\ve \theta$ can be applied to 
all neural network models $\mathcal{N}$ where the derivative $\ddthetai \, \log p_{\mathcal{N}}(\ve x^n | \ve \theta)$ in \eqref{eqn:ml-mcmc_approx} can be efficiently estimated. Since this term also has to be estimated for maximum likelihood learning \eqref{eq:delta}, synaptic sampling can basically be applied to all neuron and network models that are amenable to maximum likelihood learning. 
We illustrate 
salient new features that result from synaptic sampling (i.e., plasticity rules \eqref{eqn:ml-mcmc_approx} or \eqref{eq:onlinediscrete_mt})
for some of these
models. We begin with the
Boltzmann machine \cite{AckleyETAL:85}, one of the oldest generative neural network models. It is currently still extensively
investigated in the context of deep learning \cite{HintonETAL:06,SalakhutdinovHinton:12}.  
We demonstrate in Fig.~\ref{fig:overfitting}D,F the improved generalization capability of this model for the learning approach \eqref{equ:produces_parameter} (learning of the posterior), compared with
maximum likelihood learning (approach \eqref{eq:ml_theta}), which had been theoretically predicted by \cite{MacKay:92a} and \cite{Bishop:06}. 
But this model for learning the posterior (approach \eqref{equ:produces_parameter}) in Boltzmann machines is now based on local plasticity rules.
Note that the Boltzmann machine with synaptic sampling samples simultaneously on two different time scales: In addition to sampling for given parameters $\ve \theta$ from likely network states in the usual manner, it now samples simultaneously on a slower time scale according to \eqref{eq:onlinediscrete_mt} from the posterior of network parameters $\ve \theta$.

A Boltzmann machine 
employs extremely simple non-spiking neuron models with binary outputs. Neuron $y_i$ outputs $1$ with probability $\sigma (\sum_j w_{ij} y_j + b_i)$, else $0$, where $\sigma$ is the logistic sigmoid $\sigma(u) = \frac{1}{1+e^{-u}}$, with synaptic weights $w_{ij}$ and bias parameters $b_i$. 
Synaptic connections in a Boltzmann machine are bidirectional, with symmetric weights ($w_{ij} = w_{ji}$).
The parameters $\ve\theta$ for the Boltzmann machine consist of all weights $w_{ij}$ and biases $b_i$ in the network. 
For the special case of a restricted Boltzmann machine (RBM), maximum likelihood learning of these parameters can be done efficiently \cite{Hinton:02}, and therefore RBM's are typically used for deep learning. 
An RBM has a layered structure with one layer of visible neurons $\ve x$ and a second layer of hidden neurons $\ve z$. Synaptic connections are formed only between neurons on different layers (Fig.~\ref{fig:overfitting}A). 
The maximum likelihood gradients $\Delta w_{ij}^{ML} = \ddwij \log p_{\mathcal{N}} (\ve x | \ve \theta)$ and $\Delta b_{i}^{ML} = \ddbi \log p_{\mathcal{N}} (\ve x | \ve \theta) $ can be efficiently approximated for this model, for example
\begin{align}
 \ddwij \log p_{\mathcal{N}} (\ve x^n | \ve \theta) &\approx  \, z_i^n x_j^n - \hat{z}_i^n \hat{x}_j^n\;, \label{eq:onlinediscrete-cd-w}
\end{align}
where $x^n_j$ is the output of input neuron $j$ while input $\bx^n$ is presented, and $\hat{x}_j^n$ its output during a subsequent phase of spontaneous activity (``reconstruction phase''); analogously for the hidden neuron $z_j$ (see \nameref{sec:methods} and \nameref{sec:supplement}). 

We integrated this maximum likelihood estimate \eqref{eq:onlinediscrete-cd-w} into the synaptic sampling rule \eqref{eq:onlinediscrete_mt} in order to test
whether a suitable prior $p_\mathcal{S}(\bw)$ for the weights improves
the generalization capability of the network. 
The network received as input just five samples $\ve x^{1}, \ldots, \ve x^{5}$ of a handwritten Arabic number {\it 1} from the MNIST dataset
(the training set, shown in Fig.~\ref{fig:overfitting}A) that were repeatedly presented.
Each pixel of the digit images was represented by one neuron in the visible layer (which consisted of $784$ neurons). 
We selected a second set of 100 samples of the handwritten digit {\it 1} from the MNIST dataset as test set (Fig.~\ref{fig:overfitting}B). 
These samples include completely different styles of writing that were not present in the training set. 
After allowing the network to learn the five input samples from Fig.~\ref{fig:overfitting}A for various numbers of update steps (horizontal axis of Fig.~\ref{fig:overfitting}D,F), we evaluated the learned internal model of this network $\mathcal{N}$ for the digit $\textit{1}$ by measuring the average $\log$-likelihood $\log p_{\mathcal{N}} (\bx | \ve \theta)$ for the test data $\bx$. The result is indicated in Fig.~\ref{fig:overfitting}D,F for the training samples $\bx$ by the blue curves, and for the new test examples $\bx$, that were never shown while synaptic plasticity was active, by the red curves.
 
First,  a uniform prior over the synaptic weights was used (Fig.~\ref{fig:overfitting}C), which corresponds to the common maximum likelihood learning paradigm \eqref{eq:onlinediscrete-cd-w}. The performance on the test set (shown on vertical axis) initially increases but degrades for prolonged exposure to the training set (length of that prior exposure shown on horizontal axis). 
This effect is known as overfitting  \cite{Bishop:06,MacKay:92a}.
It can be reduced by choosing a suitable prior $p_\mathcal{S}(\ve \theta)$ in the synaptic sampling rule \eqref{eq:onlinediscrete_mt}.
The choice for the prior distribution is best if it matches the statistics of the training samples \cite{MacKay:92a}, which has in this case two modes (resulting from black and white pixels). The presence of this prior in the learning rule maintains good generalization capability for test samples even after prolonged exposure to the training set (red curve in Fig.~\ref{fig:overfitting}F).

\subsection{Spine motility as synaptic sampling}
\label{sec:spine-motility}

\begin{figure}
\centering
\includegraphics{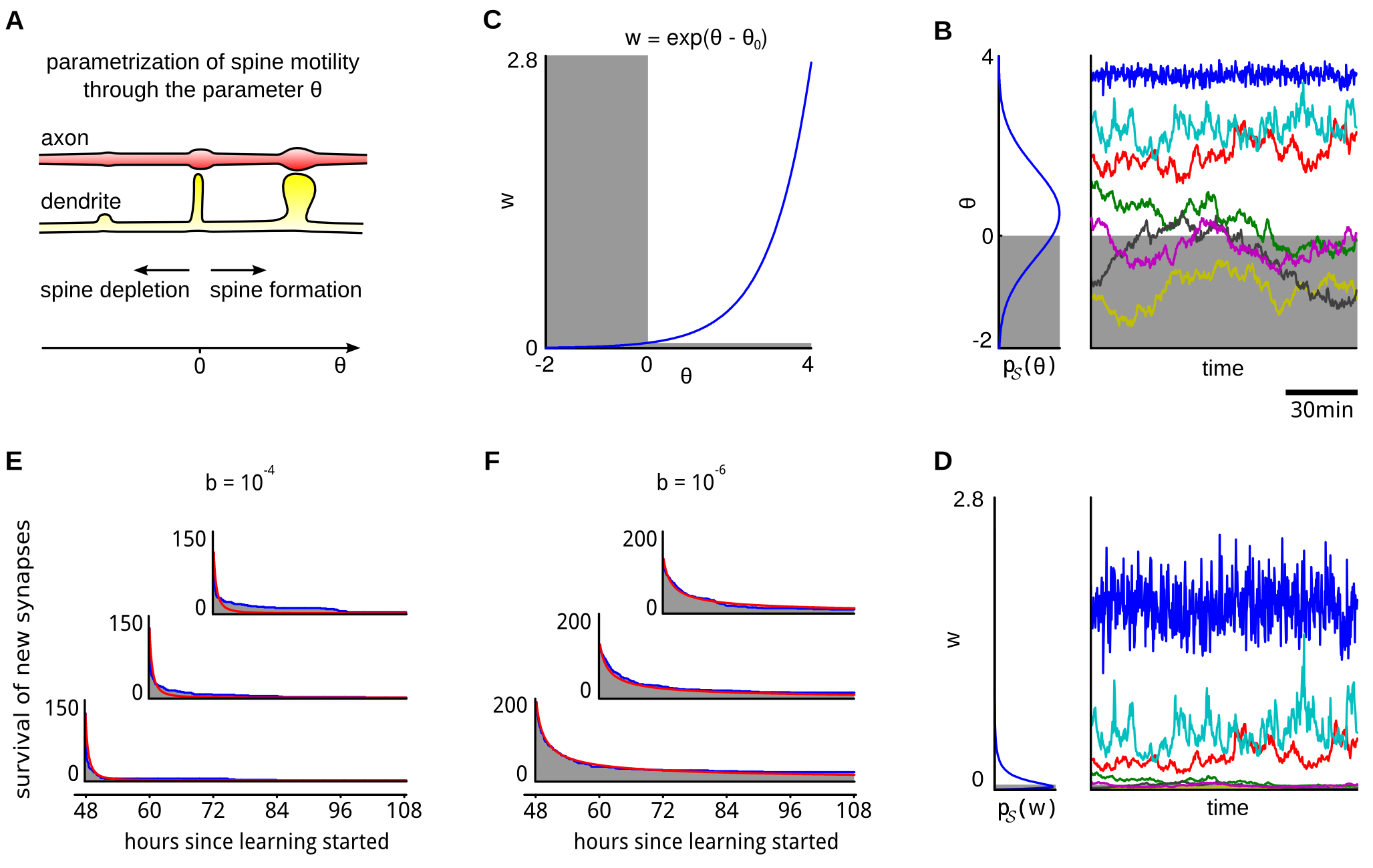}
\caption{{\bf Integration of spine motility into the synaptic sampling model.}
(A) Illustration of the parametrization of spine motility. 
Values $\theta > 0$ indicate a functional synaptic connection.
(B) A Gaussian prior $p_\mathcal{S}(\theta)$, and a few stochastic sample trajectories of $\theta$ according to the synaptic sampling rule \eqref{eq:structsamp-general}. Negative values of $\theta$ (gray area) are interpreted as non-functional connections. Some stable synaptic connections emerge (traces in the upper half), whereas other synaptic connections come and go (traces in lower half). All traces, as well as survival statistics shown in (E,F), are taken from the network simulation described in detail in the next section and Fig.~\ref{fig:structural-plasticity}. 
(C) The exponential function maps synapse  parameters $\theta$ to synaptic efficacies $w$. 
Negative values of $\theta$, corresponding to (retracted) spines are mapped to a tiny region close to zero in the $w$-space.
(D) The Gaussian prior in the $\theta$-space translates to a log-normal distribution in the $w$-space.
The traces from (B) are shown in the right panel transformed into the $w$-space. Only persistent synaptic connections contribute substantial synaptic efficacies. (E,F) The emergent survival statistics of newly formed synaptic connections, (i.e., formed during the preceding 12 hours) exhibits in our synaptic sampling model a power-law behavior (gray area denotes simulation results, red curves are power-law fits, see \nameref{sec:supplement}). The time-scale (and exponent of the power-law) depends on the learning rate $b$ in equation \eqref{eq:structsamp-general}, and can assume any value in our quite general model (shown is $b=10^{-4}$ in (E) and $b=10^{-6}$ in (F)).
}
\label{fig:prior-bars}
\end{figure}

In the following sections we apply our synaptic sampling framework to 
networks of spiking neurons and biological models for network plasticity.
The number and volume of spines for a synaptic connection is thought to be directly related to its synaptic weight \cite{LoewensteinETAL:11}.
Experimental studies have provided a wealth of information about the stochastic dynamics of dendritic spines (see e.g.~\cite{TrachtenbergETAL:02,ZuoETAL:05,HoltmaatETAL:05,
HoltmaatSvoboda:09,LoewensteinETAL:11,LoewensteinETAL:14}). 
They demonstrate that the volume of a substantial fraction of dendritic spines varies
continuously over time, and that all the time new spines and synaptic connections are formed and existing ones are eliminated.
We show that these experimental data on spine motility can be understood as special cases of synaptic sampling. The synaptic sampling framework is however very general, and many different models for spine motility can be derived from it as special cases. We demonstrate this here for one simple model, induced by the following assumptions:
\begin{enumerate}
\item We restrict ourselves to plasticity of excitatory synapses, although the framework is general enough to apply to inhibitory synapses as well.
\item In accordance with experimental studies \cite{LoewensteinETAL:11}, we require that spine sizes have a multiplicative dynamics, i.e., that the amount of change within some given time window is proportional to the current size of the spine.
\item We assume here for simplicity that a synaptic connection between two neurons is realized by a single spine and that there is a single parameter $\theta_i$ for each potential synaptic connection $i$.
\end{enumerate}
The last requirement can be met by encoding the state of the synapse in an abstract form, that  represents synaptic connectivity and synaptic plasticity in a single parameter $\theta_i$. We define that negative values of $\theta_i$ represent a current disconnection and positive values represent a functional synaptic connection. The distance of the current value of $\theta_i$ from zero indicates how likely it is that the synapse will soon reconnect (for negative values) or withdraw (for positive values), see Fig.~\ref{fig:prior-bars}A. In addition the synaptic parameter $\theta_i$ encodes for positive values the synaptic efficacy $w_i$, i.e., the resulting EPSP amplitudes, by a simple mapping $w_i=f(\theta_i)$.

A large class of mapping functions $f$ is supported by our theory (see \nameref{sec:supplement} for details). The second assumption which requires multiplicative synaptic dynamics supports an exponential function $f$ in our model, in accordance with previous models of spine motility \cite{LoewensteinETAL:11}. Thus, we assume in the following that the efficacy $w_i$ of synapse $i$ is given by
\begin{align}
 w_i = \exp( \theta_i - \theta_0 )\;,\label{eq:thetamap}
\end{align}
see Fig.~\ref{fig:prior-bars}C.
Note that for a large enough offset $\theta_0$, negative parameter values $\theta_i$ (which model a non-functional synaptic connection) are automatically mapped onto a tiny region close to zero in the $w$-space, so that retracted spines have essentially zero synaptic efficacy. 
The general rule for online synaptic sampling  \eqref{eqn:ml-mcmc_approx} for the exponential mapping \eqref{eq:thetamap} can be written as (see \nameref{sec:supplement})
\begin{align}
 d \theta_i &= b \bigg(  \ddthetai \log p_\mathcal{S}(\ve \theta) \,+\,   N w_i\,
 \ddwi \log p_{\mathcal{N}}(\ve x^n | \ve w) 
\bigg) dt \,+\,  \sqrt{2 b}\,d\wiener_i \;.\label{eq:structsamp-general}
\end{align}
In equation \eqref{eq:structsamp-general} the multiplicative synaptic dynamics becomes explicit. The gradient $\ddwi \log p_{\mathcal{N}}(\ve x^n | \ve w)$, i.e., the activity-dependent contribution to synaptic plasticity, is weighted by $w_i$. Hence, for negative values of $\theta_i$ (non-functional synaptic connection), the activities of the pre- and post-synaptic neurons have negligible impact on the dynamics of the synapse. Assuming a large enough $\theta_0$, retracted synapses therefore evolve solely according to the prior $p_\mathcal{S}(\ve \theta)$ and the random fluctuations $d\wiener_i$. For large values of $\theta_i$ the opposite is the case. The influence of the prior $\ddthetai \log p_\mathcal{S}(\ve \theta)$ and the Wiener process $d\wiener_i$ become negligible, and the dynamics is dominated by the activity-dependent likelihood term. Large synapses can therefore become quite stable if the presynaptic activity is strong and reliable (see Fig.~\ref{fig:prior-bars}B). Through the use of parameters $\ve \theta$ which determine both synaptic connectivity and synaptic efficacies, the synaptic sampling framework provides a unified model for structural and synaptic plasticity. The prior distribution
can have significant impact on the spine motility, encouraging for example sparser or denser synaptic connectivity.
If the activity-dependent second term in Eq.~\eqref{eq:structsamp-general}, that tries to maximize the likelihood, is small (e.g., because $\theta_i$ is small or parameters are near a mode of the likelihood) then Eq.~\eqref{eq:structsamp-general} implements an Ornstein Uhlenbeck process. This prediction of our model is consistent with a previous analysis which showed that an Ornstein Uhlenbeck process is a viable model for synaptic spine motility \cite{LoewensteinETAL:11}. 

The weight dynamics that emerges through the stochastic process \eqref{eq:structsamp-general} is illustrated in the right panel of Fig.~\ref{fig:prior-bars}D. 
A Gaussian parameter prior $p_{\mathcal{S}}(\theta_i)$ results in a log-normal prior $p_{\mathcal{S}}(w_i)$ in a corresponding stochastic differential equation for synaptic efficacies $w_i$ (see \nameref{sec:supplement} for details).

The last term (noise term) in our synaptic sampling rule \eqref{eq:structsamp-general} 
predicts that eliminated connections spontaneously regrow at irregular intervals. In this way they can test whether they can contribute to explaining the input. If they cannot contribute, they disappear again. 
The resulting power-law behavior of the survival of newly formed synaptic connections (Fig.~\ref{fig:prior-bars}E,F) matches corresponding new experimental data \cite{LoewensteinETAL:14} and is qualitatively similar to earlier experimental results  which revealed a quick decay of transient dendritic spines \cite{YangETAL:09,ZuoETAL:05,HoltmaatETAL:05}.
Functional consequences of this structural plasticity are explored in Fig.~\ref{fig:structural-plasticity} and \ref{fig:self-repair}.

\subsection{Fast adaptation of synaptic connections and weights to a changing input statistics}
\label{sec:adaptation}

\begin{figure}
\begin{center}
\includegraphics{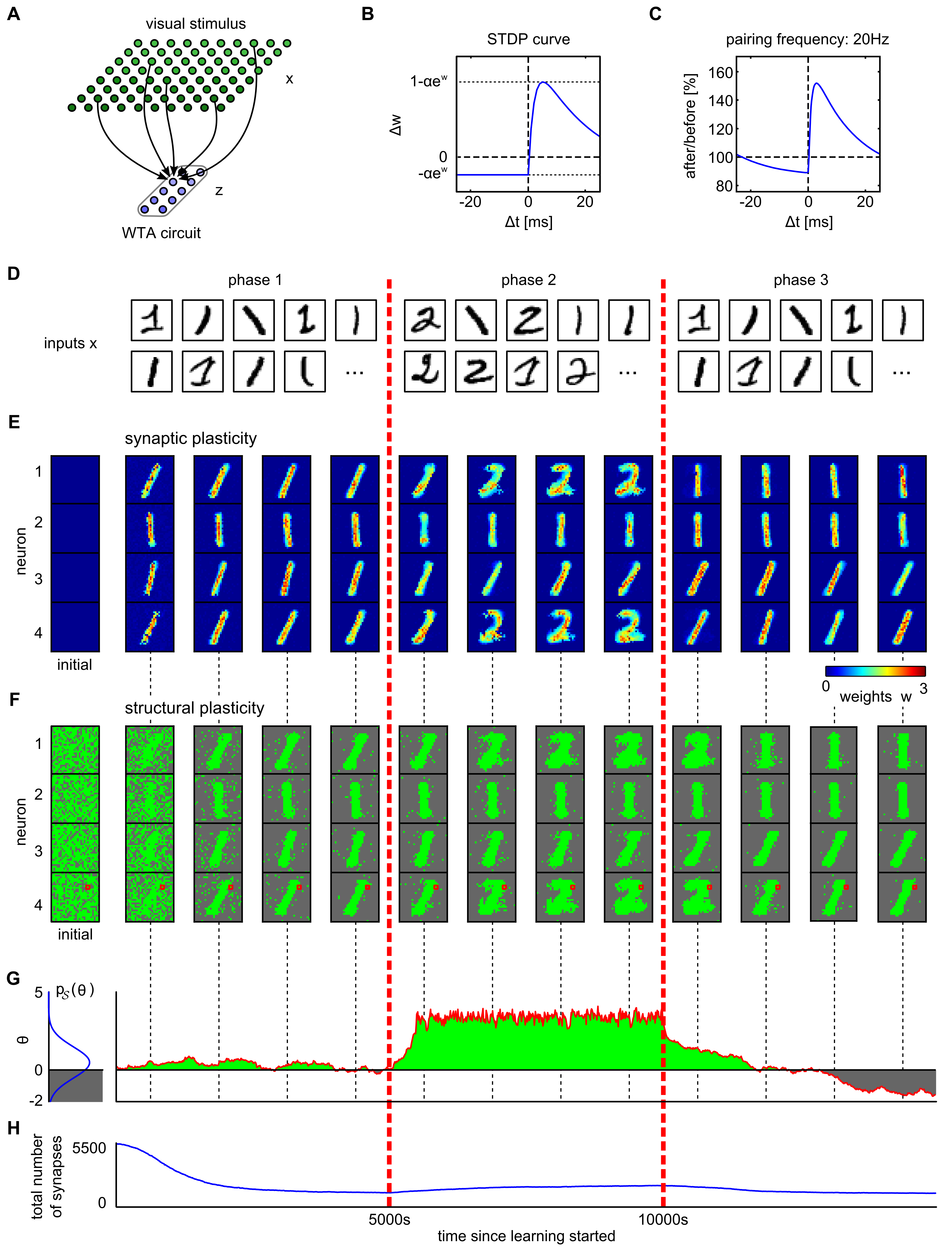}
\end{center}
\textit{Figure caption on next page...}
\end{figure}
\begin{figure}
\caption{{\bf Adaptation of synaptic connections to changing input statistics through synaptic sampling.}
(A) Illustration of the network architecture. A WTA circuit consisting of ten neurons $\ve z$ receives afferent stimuli from input neurons $\ve x$ (few connections shown for a single neuron in $\ve z$).
(B) The STDP learning curve that arises from the likelihood term in equation \eqref{eq:structsamp-spiking}. 
(C) Measured STDP curve that results from a related STDP rule for a moderate pairing frequency of 20 Hz, as in \cite{SjostromETAL:01}.  
(Figure adapted from \cite{NesslerETAL:13}).
(D) Two different data sets, both consisting of hundreds of samples of handwritten digits were presented to the network. 
In phase 1 and 3 only samples for digit 1 were shown, in phase 2 samples from digits $\textit{1}$ and $\textit{2}$.
(E,F) Evolution of synaptic parameters 
for four of the ten neurons in $\ve z$ is shown.
A snapshot of the current weight vectors (E), and currently active synaptic connections (green dots in (F)) is taken every $1000$ seconds
(i.e., after showing $4000$ further input samples).
It can be seen that most of the initial synapses are eliminated: only the connections to input neurons that encode relevant information (i.e., that are not encoding white background pixels most of the time) survive. Upon the switch to a different distribution, some eliminated synapses are reactivated, while others which were previously active, are eliminated. 
Note the constant fluctuation of functional synapses (sparse green dots in the background of (F)).
(G) Time course of a sample synaptic parameter
$\theta_{i}$ is shown (position of the synapse is indicated by red squares in (F)). The parameter prior is shown on the left.
This synaptic connection turns out to be useful for learning the digit $\textit{2}$, whereas its dynamics during the first and third phase is driven by the prior and random walk behavior (Wiener process $d \wiener$). Consequently it probes crossing of the threshold $0$ at irregular intervals during phase 1 and 3.
(H) The total number of active synapses remains low throughout learning (effect of the prior in the learning rule). 
}
\label{fig:structural-plasticity}
\end{figure}

We will explore in this and the next section implications of the synaptic sampling rule \eqref{eq:structsamp-general} for network plasticity in simple generative spike-based neural network models.

The main types of spike-based generative neural network models that have been proposed are  \cite{BreaETAL:13,BoerlinETAL:13,NesslerETAL:13,HabenschussETAL:13}. We focus here on the type of models introduced by \cite{NesslerETAL:13,HabenschussETAL:13,KappelETAL:14}, since these models allow an easy estimation of the likelihood gradient (the second term in  \eqref{eq:structsamp-general}) and can relate this likelihood term to STDP. 
Since these spike-based neural network models have non-symmetric synaptic connections
(that model chemical synapses between pyramidal cells in the cortex),
they do not allow to regenerate inputs $\ve x$ from internal responses $\ve z$ by running the network backwards (like in a Boltzmann machine). Rather they are \emph{implicit} generative models, where synaptic weights from inputs to hidden neurons are interpreted as implicit models for presynaptic activity, given that the postsynaptic neuron fires.

We focus in this section on a simple model for an ubiquitous cortical microcircuit motif: an ensemble of pyramidal cells with lateral inhibition,
often referred to as Winner-Take-All (WTA) circuit.
It has been proposed that this microcircuit motif provides for computational analysis an important bridge between single neurons and larger brain systems \cite{Carandini:12}. We employ a simple form of divisive normalization (as proposed by \cite{Carandini:12}; see \nameref{sec:methods}) 
 to model lateral inhibition, thereby arriving at a theoretically tractable version of this microcircuit motif that allows us to compute the maximum likelihood term (second term in \eqref{eq:structsamp-general}) in the synaptic sampling rule. 
We assumed Gaussian prior distributions $p_\mathcal{S}(\theta_i)$ over the synaptic parameters $\theta_i$ (as in Fig.~\ref{fig:prior-bars}B).
Then the synaptic sampling rule \eqref{eq:structsamp-general} yields for this model
\begin{align}
 d \theta_i &= b \bigg( \frac{1}{\sigma^2}\left(\theta_i - \mu \right) \;+\; N w_i\, S(t) \left(x_i(t) - \alpha\,e^{w_i}\right)
\bigg) dt \,+\,  \sqrt{2 b}\,d\wiener_i \;,\label{eq:structsamp-spiking}
\end{align}
where $S(t)$ denotes the spike train of the postsynaptic neuron and $x_i(t)$ denotes the weight-normalized value of the sum of EPSPs from presynaptic neuron $i$ at time $t$ (i.e., the summed EPSPs that would arise for weight $w_i=1$; see \nameref{sec:methods} for details). $\alpha$ is a parameter that scales the impact of synaptic plasticity in dependence on the current synaptic efficacy.
The resulting activity-dependent component $S(t)(x_i(t) - \alpha\,e^{w_i}) $ 
of the likelihood term is a simplified version of the standard STDP learning rule (Fig.~\ref{fig:structural-plasticity}B, C), like in \cite{NesslerETAL:13,KlampflMaass:13}. 
Synaptic plasticity (STDP) for connections from input neurons to pyramidal cells in the WTA circuit can be understood from the generative aspect as fitting a mixture of Poisson (or other exponential family) distributions to high-dimensional spike inputs \cite{NesslerETAL:13,HabenschussETAL:13}. 
The factor $w_i = \exp (\theta_i - \theta_0)$ had been discussed in \cite{NesslerETAL:13}, because it is compatible with the underlying generative model, but provides in addition a better fit to the experimental data of \cite{SjostromETAL:01}.

We examine in this section emergent properties of network plasticity in this simple spike-based neural network under the synaptic sampling rule \eqref{eq:structsamp-spiking}. The network was exposed to a large number of 
handwritten digits from the MNIST dataset, like in Fig.~\ref{fig:overfitting}. 
However we examined in addition network configurations that result from changes in the distribution in input patterns.
The change of the input distribution was implemented by presenting to the network in phase 1 only examples of the handwritten digit {\it 1}, then presenting in phase 2 examples of the handwritten digits {\it 1} and {\it 2}, and finally switching in phase 3 back to  examples of the digit {\it 1} only (Fig.~\ref{fig:structural-plasticity}D).
Inputs were presented sequentially and in random order, such that each sample was represented by a $200$ms-long spike pattern (see \nameref{sec:methods} for details). 
The input samples were randomly drawn from the set of all $6742$ handwritten digits {\it 1} in the MNIST database during phase $1$ and $3$. During phase $2$ the inputs were drawn from this set of handwritten {\it 1}'s combined with all $5958$ handwritten digits {\it 2} in MNIST.
The digit presentations were interleaved with $50$ms frames of random $1$Hz input activity. 
These inputs were injected into a small WTA circuit consisting of ten spiking neurons 
(Fig.~\ref{fig:structural-plasticity}A; see \nameref{sec:methods} for details). 
Synaptic sampling according to \eqref{eq:structsamp-spiking} was applied continuously to all synapses from the $784$ input neurons (encoding $28$x$28$ input pixels) to the ten neurons in the WTA circuit.

The resulting network plasticity, comprising structural plasticity and synaptic plasticity,
is presented in Fig.~\ref{fig:structural-plasticity}E-G.
The evolution of synaptic strengths is shown in Fig.~\ref{fig:structural-plasticity}E for four of the ten neurons (synaptic weights projected back into the $28$x$28$ input space for easier visualization).
Synapse elimination and regrowth can be observed in Fig.~\ref{fig:structural-plasticity}F,G. The prior acts as a constant force which tries to drive synapses towards elimination. Those synapses which have a strong positive likelihood gradient (second term in Eq.~\eqref{eq:structsamp-spiking}), indicating an important position in explaining the input, survive. The remaining synapses are eliminated as they cross the zero threshold, thereby
entering the ``potential regrowth'' regime. 
The randomly distributed isolated green dots in Fig.~\ref{fig:structural-plasticity}F show tentative regrowth of such synapses. Fig.~\ref{fig:structural-plasticity}G tracks the history of a single synaptic connection of neuron 4 (see red square in the bottom row of Fig.~\ref{fig:structural-plasticity}F). Whereas its regrowth is not supported by the likelihood term during phases 1 and 3, it becomes stable during phase 2. 
Regrowth, as opposed to elimination, is almost entirely driven by the prior and the noise term $d \wiener_i$ in \eqref{eq:structsamp-spiking}. 
This asymmetry between active and inactive synapses
follows directly from our approach to map negative parameter values onto zero synaptic efficacy: varying the parameter within the negative range has 
no effect on the likelihood term.

In contrast to preceding models for spine motility and STDP we have integrated here both into a single stochastic rule \eqref{eq:structsamp-spiking} that was derived from first principle (goal \eqref{equ:produces_parameter} and corresponding rule \eqref{eqn:ml-mcmc2}). Note that a substantial stochastic component is essential for network plasticity in order to enable a network to reconfigure in response to changes in its input distribution as in Fig.~\ref{fig:structural-plasticity}. At the same time the prior ensures that the network connections remain sparse, even in the face of challenges through a changing input distribution (see Fig.~\ref{fig:structural-plasticity}H).

\subsection{Inherent network compensation capability through synaptic sampling}
\label{sec:compensation}

\begin{figure}
\begin{center}
\includegraphics{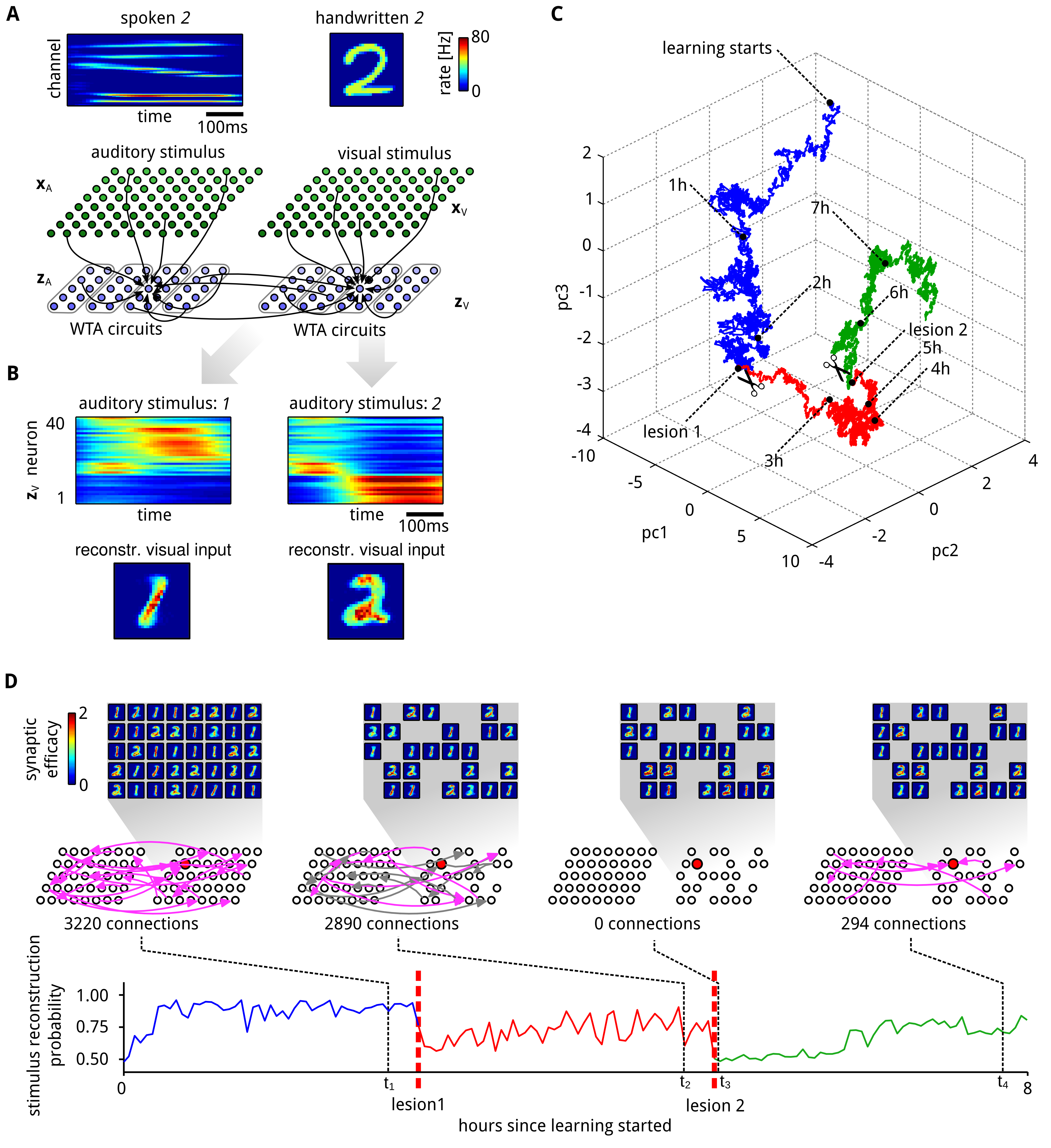}
\end{center}
\textit{Figure caption on next page...}
\end{figure}
\begin{figure}
\caption{{\bf Inherent compensation for network perturbations.}
(A) A spike-based generative neural network (illustrated at the bottom) received simultaneously spoken and handwritten representations of the same digit (and for tests only spoken digits, see (B)). Stimulus examples for spoken and written digit \textit{2} are shown at the top. 
These inputs are presented to the network through corresponding firing rates of ``auditory'' ($\bx_{A}$) and ``visual'' ($\bx_{V}$) input neurons.
Two populations $\bz_{A}$ and $\bz_{V}$  of 40 neurons, each consisting of four WTA circuits like in Fig.~\ref{fig:structural-plasticity}, receive exclusively auditory or visual inputs. In addition, arbitrary lateral excitatory connections between these ``hidden'' neurons are allowed.
(B) Assemblies of hidden neurons emerge that encode the presented digit (\textit{1} or \textit{2}). Top panel shows PETH of all neurons from $\bz_{V}$ for stimulus \textit{1} (left) and \textit{2} (right) after learning, when only an auditory stimulus is presented. Neurons are sorted by the time of their highest average firing. Although only auditory stimuli are presented, it is possible to reconstruct 
an internally generated ``guessed'' visual stimulus that represents the same digit (bottom). 
(C) 
First three PCA components of the temporal evolution of a subset $\bm \theta'$ of network parameters $\bm \theta$ (see text).
Two major lesions were applied to the network. In the first lesion (transition to red) all neurons that significantly encode stimulus \textit{2} were removed from the population $\bz_{V}$. In the second lesion (transition to green) all currently existing
synaptic connections between neuron in $\bz_{A}$ and $\bz_{V}$ were removed, and not allowed to regrow. After each lesion the network parameters $\bm \theta'$ migrate to a new manifold. 
(D) The generative reconstruction performance of the ``visual'' neurons $\bz_{V}$ for the test case when only an auditory stimulus is presented was tracked throughout the whole learning session, including lesions \textit{1} and \textit{2} (bottom panel). 
After each lesion the performance strongly degrades, but reliably recovers.
Insets show at the top the synaptic weights 
of neurons in $\bz_{V}$ at $4$ time points $t_1, \ldots, t_4$, projected back into the input space like in Fig.~\ref{fig:structural-plasticity}E. 
Network diagrams 
in the middle show ongoing network rewiring for synaptic connections between the hidden neurons $\bz_{A}$ and $\bz_{V}$. Each arrow indicates a functional connection between two neurons. To keep the figure uncluttered only subsets of synapses are shown (1\% randomly drawn from the total set of possible lateral connections). 
Connections at time $t_2$ that were already functional at time $t_1$ are plotted in gray. The neuron whose parameter vector $\bm \theta'$ is tracked in (C) is highlighted in red. The text under the network diagrams shows the total number of functional connections between hidden neurons at the time point.
}
\label{fig:self-repair}
\end{figure}


Numerous experimental data show that the same function of a neural circuit is achieved in different individuals with drastically different parameters, and also that a single organism can compensate for disturbances by moving to a new parameter vector \cite{TangETAL:10,GrashowETAL:10,MarderTaylor:11,Marder:11,PrinzETAL:04}. 
These results suggest that there exists
some low-dimensional submanifold of values for the high-dimensional parameter vector $\ve \theta$ of a biological neural network that all provide stable network function (degeneracy). 
We propose that the previously discussed posterior distribution of network parameters $\ve \theta$ provides a mathematical model for such low-dimensional submanifold. Furthermore we propose that the underlying continuous stochastic fluctuation $d\wiener$ provides a driving force that automatically moves network parameters (with high probability) to a functionally more attractive regime when the current solution becomes less performant because of perturbations, such as lesions of neurons or network connections.
This compensation capability is not an add-on to the synaptic sampling model, but an inherent feature of its organization.

We demonstrate this inherent compensation capability in Fig.~\ref{fig:self-repair} for a 
generative spiking neural network  
with synaptic parameters $\ve \theta$ that regulate simultaneously structural plasticity and synaptic plasticity (dynamics of weights) as in Fig.~\ref{fig:prior-bars} and \ref{fig:structural-plasticity}. 
The prior $p_\mathcal{S}(\ve \theta)$ for these parameters is here the same as in the preceding section (see Fig.~\ref{fig:structural-plasticity}G on the left). But in contrast to the previous section we consider here a network that allows us to study the self-organization of connections \emph{between} hidden neurons.  The network consists 
of eight WTA-circuits, but in contrast to Fig.~\ref{fig:structural-plasticity} we allow here arbitrary excitatory synaptic connections between neurons within the same or different ones of these WTA circuits. This network models multi-modal sensory integration and association in a simplified manner. Two populations of ``auditory'' and ``visual'' input neurons $\bx_A$ and $\bx_V$ project onto corresponding populations $\bz_A$ and $\bz_V$ of hidden neurons (each consisting of one half of the WTA circuits, see lower panel of Fig.~\ref{fig:self-repair}A). Only a fraction of the potential synaptic connections became functional (see supplementary Fig.~S1A) through the synaptic sampling rule \eqref{eq:structsamp-spiking} that integrates structural and synaptic plasticity. Synaptic weights and connections were not forced to be symmetric or bidirectional.

As in the previous demonstrations we do not use external rewards or teacher-inputs for guiding  network plasticity. Rather, we allow the model to discover on its own regularities in its network inputs. 
The ``auditory" hidden neurons $\bz_A$ on the left in Fig.~\ref{fig:self-repair}A received temporal spike patterns
from the auditory input neurons $\bx_A$
that were generated from spoken utterings of the digit \textit{1} and \textit{2}
(which lasted between $320$ and $520$ms).
Simultaneously we presented to the ``visual" hidden neurons $\bz_V$ on the right for the same time period a (symbolic) visual representation of the same digit (randomly drawn from the MNIST database like in Fig.~\ref{fig:overfitting} and \ref{fig:structural-plasticity}). 

The emergent associations between 
the two populations $\bz_A$ and $\bz_V$ of hidden neurons were
tested by presenting auditory input only 
and observing the activity
of the ``visual'' hidden neurons $\bz_V$.
Fig.~\ref{fig:self-repair}B shows the emergent activity of the neurons $\bz_V$ when only the auditory stimulus was presented (visual input neurons $\bx_V$ remained silent). 
The generative aspect of the network can be demonstrated by reconstructing
for this case the visual stimulus from the 
activity of the ``visual'' hidden neurons $\bz_V$.
Fig.~\ref{fig:self-repair}B shows reconstructed visual stimuli from a single run where only the auditory stimuli $\bx_A$ for digits  \textit{1} (left) and \textit{2} (right) were presented to the network. Digit images were reconstructed by multiplying the synaptic efficacies of synapses from neurons in $\bx_V$ to neurons in $\bz_V$ (which did not receive any input from $\bx_V$ in this experiment) with the instantaneous firing rates of the corresponding $\bz_V$-neurons.

In order to investigate the inherent compensation capability of synaptic sampling, we applied two lesions to the network within a learning session of more than $7$ hours.  In the first lesion all neurons (16 out of 40) that became tuned for digit \textit{2} in the preceding learning (see Fig.~\ref{fig:self-repair}D and \nameref{sec:supplement}) were removed. 
The lesion significantly impaired the performance of the network in stimulus reconstruction, but it was able to recover from the lesion after about one hour of continuing network plasticity according to Eq.~\eqref{eq:structsamp-spiking} (Fig.~\ref{fig:self-repair}D). The reconstruction performance of the network was measured here continuously through the capability of a linear readout neuron from the visual ensemble to classify the current auditory stimulus (\textit{1} or \textit{2}).

In the second lesion all synaptic connections between hidden neurons that were present after recovery from the first lesion
were removed and not allowed to regrow (2936 synapses in total).  
After about two hours of continuing synaptic sampling $294$ new synaptic connections between hidden neurons emerged.
These made it again possible to infer the auditory stimulus from the activity of the remaining $24$ hidden neurons in the population $\bz_V$ (in the absence of any input from the population $\bx_V$), at about 75\% of the performance level before the second lesion (see bottom panel of Fig.~\ref{fig:self-repair}D).  

In order to illustrate the ongoing network reconfiguration we track in Fig.~\ref{fig:self-repair}C the temporal evolution of a subset $\bm \theta'$ of network parameters (35 parameters $\theta_i$ associated with the potential synaptic connections of the neuron marked in red in the middle of Fig.~\ref{fig:self-repair}D from or to other hidden neurons, excluding those that were removed at lesion \textit{2} and not allowed to regrow).
The first three PCA components of this $35$-dimensional parameter vector are shown.
The vector $\bm \theta'$ fluctuates first within one region of the parameter space while probing different solutions to the learning problem, e.g., high probability regions of the posterior distribution (blue trace). Each lesions induced a fast switch to a different region (red,green), accompanied by a recovery of the visual stimulus reconstruction performance (see Fig.~\ref{fig:self-repair}D). 

Altogether this experiment showed that continuously ongoing synaptic sampling maintains stable network function also in a more complex network architecture. Another consequence of synaptic sampling was that the neural codes (assembly sequences) that emerged for the two digit classes within the hidden neurons $\bz_A$ and $\bz_V$ (see supplementary Fig.~S1B) drifted over larger periods of time (also in the absence of lesions), similarly as observed for place cells in \cite{ZivETAL:13} and for tuning curves of motor cortex neurons in \cite{RokniETAL:07}.

\section{Discussion}

We have shown that stochasticity may provide an important function for network plasticity.
It enables networks to sample parameters from some low-dimensional manifold in a high-dimensional parameter space that represents attractive combinations of structural constraints and rules (such as sparse connectivity and heavy-tailed distributions of synaptic weights) and a good fit to empirical evidence (e.g., sensory inputs).
We have developed a normative model for this new learning perspective, where the traditional gold standard of maximum likelihood optimization is replaced by theoretically optimal sampling from a posterior distribution of parameter settings, where regions of high probability provide a theoretically optimal model for the low-dimensional manifold from which parameter settings should be sampled. The postulate that networks should learn such posterior distributions of parameters, rather than maximum likelihood values, had been 
proposed already for quite some while for artificial neural networks \cite{MacKay:92a,Bishop:06}, since such organization of learning promises better generalization capability to new examples. The open problem how such posterior distributions could be learned by networks of neurons in the brain, in a way that is consistent with experimental data, has been highlighted in \cite{PougetETAL:13} as a key challenge for computational neuroscience. We have presented here such a model, whose primary innovation is to view experimentally found trial-to-trial variability and ongoing fluctuations of parameters such as spine volumes
no longer as 
a nuisance, but as a functionally important component of the organization of network learning, since it enables sampling from a distribution of network configurations. 
The mathematical framework that we have presented provides a normative model for evaluating such empirically found stochastic dynamics of network parameters, and for relating specific properties of this ``noise'' to functional aspects of network learning.

Reports of trial-to-trial variability and ongoing fluctuations of parameters related to synaptic weights are ubiquitous in experimental studies of synaptic plasticity and its molecular implementation, from 
fluctuations of proteins  such as PSD-95 \cite{GrayETAL:06} in the postsynaptic density that are thought to be related to synaptic strength, over intrinsic fluctuations in spine volumes and synaptic connections \cite{YasumatsuETAL:08,HoltmaatETAL:05,StettlerETAL:06,YamahachiETAL:09,HoltmaatSvoboda:09,LoewensteinETAL:11,LoewensteinETAL:14}, to surprising shifts of neural codes on a larger time scale \cite{RokniETAL:07,ZivETAL:13}. These fluctuations may have numerous causes, from noise in the external environment over noise and fluctuations of internal states in sensory neurons and brain networks, to noise in the pre- and postsynaptic molecular machinery that implements changes in synaptic efficacies on various time scales \cite{RibraultETAL:11}. One might even hypothesize, that it would be very hard for this molecular machinery to implement synaptic weights that remain constant in the absence of learning, and deterministic rules for synaptic plasticity,
because the half-life of many key proteins that are involved is relatively short, and receptors and other membrane-bound proteins are subject to Brownian motion. 
In this context the finding that neural codes shift over time \cite{RokniETAL:07,ZivETAL:13} appears to be less surprising. 
In fact, our model predicts (see \nameref{sec:supplement}) that also stereotypical assembly sequences that emerge in our model through learning, similarly as in the experimental data of \cite{HarveyETAL:12}, are subject to such shifts on a larger time scale.
However it should be pointed out that our model is agnostic with regard to the time scale on which these changes occur, since this time scale can be defined arbitrarily through the parameter $b$ (learning rate) in Eq.~\eqref{eqn:ml-mcmc2}. 

The model that we have presented makes no assumptions about the actual sources of noise. It only assumes that salient network parameters are subject to stochastic processes, that are qualitatively similar to those which have been studied and modeled in the context of Brownian motion of particles as random walk on the microscale. One can scale the influence of these stochastic forces in the model by a parameter 
$T$ that regulates the ``temperature'' of the stochastic dynamics of network parameters $\ve \theta$. 
This parameter $T$ regulates the tradeoff between trying out different regions (or modes) of the posterior distribution of $\ve \theta$ (exploration), and staying for longer time periods in a high probability region of the posterior (exploitation).  
We conjecture that this parameter $T$ varies in the brain between different brain regions, and possibly also between different types of synaptic connections within a cortical column. 
For example, spine turnover is increased for large values of $T$, and network parameters $\ve \theta$ can move faster to a new peak in the posterior distribution, thereby supporting faster learning (and faster forgetting). Since spine turnover is reported to be higher in the hippocampus than in the cortex \cite{AttardoETAL:14}, such higher value of $T$ could for example be more adequate for modeling network plasticity in the hippocampus. This model would then also support the hypothesis of  \cite{AttardoETAL:14} that memories are more transient in the hippocampus. 
In addition $T$ is likely to be regulated on a larger time scale by developmental processes, and on a shorter time scale by neuromodulators and attentional control.
The view that synaptic plasticity is stochastic had already been explored through simulation studies in 
\cite{RokniETAL:07,AjemianETAL:13}. 
Artificial neural networks were trained in \cite{AjemianETAL:13} through supervised learning with high learning rates and high amounts of noise both on neuron outputs and synaptic weight changes. 
The authors explored the influence of various combinations of noise levels and learning rates on the success of learning, which can be understood as varying the temperature parameters $T$ in the synaptic sampling framework.  
In order to measure this parameter $T$ experimentally in a direct manner, one would have to apply repeatedly the same plasticity induction protocol to the same synapse, with a complete reset of the internal state of the synapse between trials, and measure the resulting trial-to-trial variability of changes of its synaptic efficacy. Since such complete reset of a synaptic state appears to be impossible at present, one can only try to approximate it by the variability that can be measured between different instances of the same type of synaptic connection.

We have shown that the Fokker-Planck equation, a standard tool in physics for analyzing the temporal evolution of the spatial probability density function for particles under Brownian motion, can be used to create bridges between details of local stochastic plasticity processes on the microscale and the probability distribution of the vector $\ve \theta$ of all parameters on the network level. This theoretical result provides the basis for the new theory of network plasticity that we are proposing. In particular, this link allows us to derive rules for synaptic plasticity which enable the network to learn, and represent in a stochastic manner, a desirable posterior distribution of network parameters; in other words: to approximate Bayesian inference. 

We find that resulting rules for synaptic plasticity contain the familiar term for maximum likelihood learning. But another new term, apart from the Brownian-motion-like stochastic term, is the 
term $\dd{\theta_i} \log\, p_\mathcal{S}(\theta_i)$ that results from a prior distributions $p_\mathcal{S}(\theta_i)$, which could actually be different for each biological parameter $\theta_i$
and enforce structural requirements and preferences of networks of neurons in the brain.
Some systematic dependencies of changes in synaptic weights (for the same pairing of pre- and postsynaptic activity) on their current values had already been reported in \cite{LiaoETAL:92,BiPoo:98,SjostromETAL:01,MontgomeryETAL:01}. These can be modeled as impact of priors.
Other potential functional benefits of priors (on emergent selectivity of neurons) have recently been demonstrated in \cite{XiongETAL:14} for a restricted Boltzmann machine. An interesting open question is whether the non-local learning rules of their model can be approximated through biologically more realistic local plasticity rules, e.g. through synaptic sampling.
We also have demonstrated in Fig.~\ref{fig:prior-bars} that suitable priors can model experimental data from \cite{LoewensteinETAL:14} on the survival statistics of 
dendritic spines. Finally, we have demonstrated in Fig.~\ref{fig:structural-plasticity} and \ref{fig:self-repair} that suitable priors for network parameters $\theta_i$ that model spine volumes endow a neural network with the capability to respond to changes in the input distribution and network perturbations with a network rewiring that maintains or restores the network function, while simultaneously observing structural constraints such as sparse connectivity. 

Our model underlines the importance of further experimental investigation of priors for network parameters.
How are they implemented on a molecular level? What role does gene regulation have in their implementation?
How does the history of a synapse affect its prior? In particular, can consolidation of a synaptic weight $\theta_i$ be modeled in an adequate manner as a modification of its prior? This would be attractive from a functional perspective, because according to our model it both allows long-term storage of learned information and flexible network responses to subsequent perturbations.

We have focused in the examples for our model on the plasticity of synaptic weights and synaptic connections. But the synaptic sampling framework can also be used for studying the plasticity of other synaptic parameters, e.g., parameters that control the short term dynamics of synapses, i.e., their individual mixture of short term facilitation and depression. The corresponding parameters $U, D, F$ of the models from \cite{VarelaETAL:97,MarkramETAL:98} are known to depend in a systematic manner on the type of pre- and postsynaptic neuron \cite{MarkramETAL:04}, indicative of a corresponding prior. However also a substantial variability within the same type of synaptic connections, had been found \cite{MarkramETAL:04}. Hence it would be interesting to investigate functional properties and experimentally testable consequences of stochastic plasticity rules of type \eqref{eqn:ml-mcmc_approx} for $U, D, F$, and to compare the results with those of previously considered deterministic plasticity rules for $U, D, F$ (see e.g.,~\cite{NatschlaegerETAL:01a}). 

We have demonstrated that our framework
provides a new and principled way of modeling 
structural plasticity \cite{May:11,CaroniETAL:12}.
The challenge to find a biologically plausible way of modeling structural plasticity as Bayesian inference has been highlighted by \cite{PougetETAL:13}. 
One nice feature of our approach is that network rewiring and changes of synaptic weights are modeled by a single rule, that can be directly related to functional aspects of the network via the resulting posterior distribution. 
We have shown in Fig.~\ref{fig:prior-bars} and \ref{fig:structural-plasticity} that this rule produces a population of persistent synapses that remain stable over long periods of time, and another population of transient synaptic connections which disappear and reappear randomly, thereby supporting automatic adaptation of the network structure to changes in the distribution of external inputs (Fig.~\ref{fig:structural-plasticity}) and network perturbation (Fig.~\ref{fig:self-repair}).

On a more general level we propose that a framework for network plasticity where network parameters are sampled continuously from a posterior distribution 
will automatically be less brittle than 
previously considered maximum likelihood learning frameworks. The latter require some intelligent supervisor who recognizes that the solution given by the current parameter vector is no longer useful, and induces the network to resume plasticity. 
In contrast, plasticity processes remain active all the time in our sampling-based framework. 
Hence network compensation for external or internal perturbations is automatic and inherent in the organization of network plasticity.

The need to rethink observed parameter values and plasticity processes in biological networks of neurons in a way which takes into account their astounding variability and compensation capabilities has been emphasized by Eve Marder (see e.g. \cite{PrinzETAL:04,MarderGoaillard:06,Marder:11}) and others.
This article has introduced a new conceptual and mathematical framework for network plasticity that promises to provide a foundation for such rethinking of network plasticity.

\section{Methods}
\label{sec:methods}

\subsection{Details to {\em Learning a posterior distribution through stochastic synaptic plasticity}}\label{sec:meth_thm1}

Here we prove that $p^*(\ve \theta) = p(\ve \theta| \,\bx)$ is the unique stationary distribution of the parameter dynamics \eqref{eqn:ml-mcmc2} that operate on the network parameters $\ve \theta = (\theta_1, \dots, \theta_M)$. 
Convergence to this stationary distribution then follows for strictly positive $p^*(\ve \theta)$.
In fact, we prove here a more general result for parameter dynamics given by
\begin{align}
 d \theta_i \;=\; \left( b(\theta_i) \ddthetai \,\log p_\mathcal{S}(\ve \theta) + b(\theta_i) \ddthetai \log p_{\mathcal{N}} (\bx | \ve \theta) + T\,b'(\theta_i) \right)\; dt  \; +\sqrt{2 T b(\theta_i)} \, d \wiener_i 
 \label{eqn:ml-mcmc2-meth}
\end{align}
(for $i=1, \dots, M$). This dynamics 
includes a temperature parameter T and a sampling-speed factor $b(\theta_i)$ that can in general depend on the current value of the parameter $\theta_i$. The temperature parameter $T$ can be used to scale the diffusion term (i.e., the noise). The sampling-speed factor controls the speed of sampling, i.e., how fast the parameter space is explored. It can be made dependent on the individual parameter value without changing the stationary distribution. For example, the sampling speed of a synaptic weight can be slowed down if it reaches very high or very low values.
Note that the dynamics \eqref{eqn:ml-mcmc2} is a special case of the dynamics \eqref{eqn:ml-mcmc2-meth} with unit temperature $T=1$ and constant sampling speed $b(\theta_i)\equiv b$.
We show that the stochastic dynamics \eqref{eqn:ml-mcmc2-meth} leaves the distribution 
\begin{align}
p^{*}(\ve \theta) \equiv  \frac{1}{\mathcal{Z}} q^*(\ve \theta) 
 \end{align}
invariant, 
where $\mathcal{Z}$ is a normalizing constant $\mathcal{Z} = \int q^*(\ve \theta) \, d \ve \theta$ and
\begin{align}
 q^*(\ve \theta)  \;=\; p( \ve \theta \,|\,\bx)^\frac{1}{T}\;. \label{eqn:stat-dist-no-norm}
\end{align}
Note that the stationary distribution $p^{*}(\ve \theta)$ is shaped by the temperature parameter $T$, in the sense that $p^{*}(\ve \theta)$ is a flattened version of the posterior for high  temperature.
The result is formalized in the following theorem, which is proven in detail in \nameref{sec:supplement}:
 
\begin{thm}
\label{lem:single}
Let $p(\bx, \ve \theta)$ be a strictly positive, continuous probability distribution over continuous or discrete states $\ve x^n$ and continuous parameters $\ve \theta = (\theta_1, \dots, \theta_M)$, twice continuously differentiable with respect to $\ve \theta$. Let $b(\theta)$ be a strictly positive, twice continuously differentiable function. Then
the set of stochastic differential equations \eqref{eqn:ml-mcmc2-meth}
leaves the distribution $p^{*}(\ve \theta)$ invariant.
Furthermore, $p^{*}(\ve \theta)$ is the unique stationary distribution of the sampling dynamics.
\end{thm}

\subsubsection{Online approximation}

We show here that the rule \eqref{eqn:ml-mcmc_approx} is a
reasonable approximation to the batch-rule \eqref{eqn:ml-mcmc2}. According to the dynamics \eqref{eqn:ml-mcmc2-meth}, synaptic plasticity rules that implement synaptic sampling have to compute the log likelihood derivative $\ddthetai \log p_{\mathcal{N}} (\bx | \ve \theta)$.
We assume that every $\tau_x$ time units a different input $\ve x^{n}$ is presented to the network. For simplicity, assume that $\ve x^1, \dots, \ve x^N$ are visited in a fixed regular order. Under the assumption that input patterns are drawn independently, the likelihood of the generative model factorizes
\begin{equation}\label{eq:factorized_joint-methods}
p_{\mathcal{N}} (\bx, | \ve \theta) =  \prod^N_{n=1} p_{\mathcal{N}} (\ve x^n | \ve \theta). 
\end{equation}
The derivative of the log likelihood is then given by
\begin{align} \label{eq:dtheta_batch}
 \ddthetai \log p_{\mathcal{N}}(\bx | \ve \theta) = \sum_{n=1}^N \ddthetai \log p_{\mathcal{N}}(\ve x^{n} | \ve \theta)\;.
\end{align}
Using Eq.~\eqref{eq:dtheta_batch} in the dynamics \eqref{eqn:ml-mcmc2-meth}, one obtains
\begin{align}
 d \theta_i \;=\; \left( b(\theta_i) \ddthetai \,\log p_\mathcal{S}(\ve \theta) + b(\theta_i) \sum_{n=1}^N \ddthetai \,\log p_{\mathcal{N}}(\ve x^{n} | \ve \theta)
  + T\,b'(\theta_i) \right)\; dt  \; +\sqrt{2 T b(\theta_i)} \, d \wiener_i .
 \label{eqn:dyn_batch}
\end{align}
Hence, the parameter dynamics depends at any time on all network inputs and network responses.

This ``batch'' dynamics does not map readily onto a network implementation because the weight update requires at any time knowledge of all inputs $\ve x^{n}$. We provide here an online approximation for small sampling speeds. 
To obtain an online learning rule, we consider the parameter dynamics
\begin{align}
 d \theta_i \;=\; \left( b(\theta_i) \ddthetai \,\log p_\mathcal{S}(\ve \theta) + N b(\theta_i) \ddthetai \log p_{\mathcal{N}}(\ve x^{n} | \ve \theta)
  + T\,b'(\theta_i) \right)\; dt  \; +\sqrt{2 T b(\theta_i)} \, d \wiener_i .
 \label{eqn:dyn_online}
\end{align}
As in the batch learning setting, we assume that each input $\ve x^n$ is presented for a time interval of $\tau_x$. Integrating the parameter changes \eqref{eqn:dyn_online} over one full presentation of the data $\bx$, i.e., starting from $t=0$ with some initial parameter values $\ve \theta(0)$ up to time $t=N \tau_x$, we obtain for slow sampling speeds ($N\tau_x b(\theta_i) \ll 1$)
\begin{align}
 \theta_i(N\tau_x) - \theta_i(0) \;\approx \; N \tau_x \left( b(\theta_i) \ddthetai \,\log p_\mathcal{S}(\ve \theta) + b(\theta_i) \sum_{n=1}^N \ddthetai \log p_{\mathcal{N}}(\ve x^{n} | \ve \theta)
  + T\,b'(\theta_i) \right)  \; \nonumber \\ +\sqrt{2 T b(\theta_i)} \,  (\wiener_i^{N \tau_x} - \wiener_i^0) .
 \label{eqn:parameter_integral}
\end{align}
This is also what one obtains when integrating Eq.~\eqref{eqn:dyn_batch} for $N\tau_x$ time units (for slow $b(\theta_i)$). Hence, for slow enough $b(\theta_i)$, Eq.~\eqref{eqn:dyn_online} is a good approximation of optimal weight sampling. The update rule \eqref{eqn:ml-mcmc_approx} follows from \eqref{eqn:dyn_online} for $T=1$ and $b(\theta_i)\equiv b$.

\subsubsection{Discrete time approximation}

Here we provide the derivation for the approximate discrete time learning rule \eqref{eq:onlinediscrete_mt}.
For a discrete time parameter update at time $t$ with discrete time step $\Delta t$ during which $\ve x^n$ is presented,
a corresponding rule can be obtained by short integration of the continuous time rule \eqref{eqn:dyn_online} over the time interval from $t$ to $t+\Delta t$:
\begin{align}
 \Delta \theta_i \;&= \; \Delta t \left( b(\theta_i) \ddthetai \,\log p_\mathcal{S}(\ve \theta) + N b(\theta_i) \ddthetai \log p_{\mathcal{N}}(\ve x^{n} | \ve \theta)
  + T\,b'(\theta_i) \right)  \; 
  +\sqrt{2 T b(\theta_i)} \,  (\wiener_i^{t+\Delta t} - \wiener_i^t) \\
  &= \; \Delta t \left( b(\theta_i) \ddthetai \,\log p_\mathcal{S}(\ve \theta) + N b(\theta_i) \ddthetai \log p_{\mathcal{N}}(\ve x^{n} | \ve \theta)
  + T\,b'(\theta_i) \right)  \; 
  +\sqrt{2 T \Delta t \, b(\theta_i)} \, \nu_i^t ,
 \label{eq:rule_discrete}
\end{align}
where $\nu_i^t$ denotes Gaussian noise $\nu_i^t \sim \Nd(0, 1)$. The update rule \eqref{eq:onlinediscrete_mt} is obtained by choosing a constant $b(\theta)\equiv b$, $T=1$, and defining $\eta=\Delta t \, b$.

\subsubsection{Synaptic sampling with hidden states}

When there is a direct relationship between network parameters $\ve \theta$ and the distribution over input patterns $\ve x^n$, the parameter dynamics can directly be derived from the derivative of the data log likelihood and the derivative of the parameter prior. Typically however, generative models for brain computation assume that the network response $\ve z^n$ to input pattern $\ve x^n$ represents in some manner the value of hidden variables that explain current input pattern. In the presence of hidden variables, maximum likelihood learning cannot be applied directly, since the state of the hidden variables is not known from the observed data. The expectation maximization algorithm \cite{Bishop:06} can be used to overcome this problem. We adopt this approach here. In the online setting, when pattern $\ve x^n$ is applied to the network, it responds with network state $\ve z^n$ according to $p_{\mathcal N}(\ve z|\ve x^n, \ve \theta)$, where the current network parameters are used in this inference process. The parameters are updated in parallel according to the dynamics
\begin{align}
 d \theta_i \;=\; \left( b(\theta_i) \ddthetai \,\log p_\mathcal{S}(\ve \theta) + N b(\theta_i) \ddthetai \log p_{\mathcal{N}}(\ve x^{n}, \ve z^{n} | \ve \theta)
  + T\,b'(\theta_i) \right)\; dt  \; +\sqrt{2 T b(\theta_i)} \, d \wiener_i .
 \label{eqn:dyn_online_hidden}
\end{align}
Note that in comparison with the dynamics \eqref{eqn:dyn_online}, the likelihood term now also contains the current network response $\ve z^n$. It can be shown that this dynamics leaves the stationary distribution
\begin{align}
p^{*}(\ve \theta) \equiv  \frac{1}{\mathcal{Z}} p( \ve \theta \,|\,\bx, \bz)^\frac{1}{T},
 \end{align}
invariant, where $\mathcal{Z}$ is again a normalizing constant (the dynamics \eqref{eqn:dyn_online_hidden} is again the online-approximation). Hence, in this setup, the network samples concurrently from circuit states (given $\ve \theta)$ and network parameters (given the network state $\ve z^n$), which can be seen as a sampling-based version of online expectation maximization.

\subsection{Details to \textit{Improving the generalization capability of a neural network through synaptic sampling (Fig.~\ref{fig:overfitting}})}\label{sec:meth_generalization}

For learning the distribution over different writings of digit {\it 1} with different priors in Fig.~\ref{fig:overfitting}, a restricted Boltzmann machine (RBM) with $748$ visible and $9$ hidden neurons was used. A detailed definition of the RBM model and additional details to the simulations are given in \nameref{sec:supplement}.

\subsubsection{Network inputs}
Handwritten digit images were taken from the MNIST dataset \cite{LeCunETAL:98}. In MNIST, each instance of a handwritten digit is represented by a 784-dimensional vector $\ve x^n$. Each entry is given by the gray-scale value of a pixel in the $28 \times 28$ pixel image of the handwritten digit. The pixel values were scaled to the interval $[0, 1]$. In the RBM, each pixel was represented by a single visible neuron.
When an input was presented to the network, the output of a visible neuron was set to $1$ with probability as given by the scaled gray-scale value of the corresponding pixel.

\subsubsection{Learning procedure}
In each parameter update step the contrastive divergence algorithm  of \cite{Hinton:02} was used to estimate the likelihood gradients. Therefore, each update step consisted of a ``wake'' phase, a ``reconstruction'' phase, and the update of the parameters. The ``wake'' samples were generated by setting the outputs of the visible neurons to the values of a randomly chosen digit $\ve x^n$ from the training set and drawing the outputs $z^n_i$ of all hidden layer neurons for the given visible output. The ``reconstruction'' activities $\hat{x}^n_j$ and $\hat{z}^n_i$ were generated by starting from this state of the hidden neurons and then drawing outputs of all visible neurons. After that, the hidden neurons were again updated and so on. In this way we performed five cycles of alternating visible and hidden neuron updates. The outputs of the network neurons after the fifth cycle were taken as the resulting ``reconstruction'' samples $\hat{x}_j^n$ and $\hat{z}_i^n$ and used for the parameter updates \eqref{eq:onlinediscrete-bm-bhid}--\eqref{eq:onlinediscrete-bm-w} given below. This update of parameters concluded one update step.  

Log likelihood derivatives for the biases $b_i^{\text{hid}}$ of hidden neurons are approximated in the contrastive divergence algorithm \cite{Hinton:02} as $\frac{\partial}{\partial b_i^{\text{hid}}} \log p_{\mathcal{N}} (\ve x^n | \theta) \approx z_i^n - \hat{z}_i^n$ (the derivatives for visible biases $b_j^{\text{vis}}$ are analogous). Using Eq.~\eqref{eq:onlinediscrete_mt}, the synaptic sampling update rules for the biases are thus given by
\begin{align}
 \Delta b_i^{\text{hid}} &= \eta \, N\left(z_i^n - \hat{z}_i^n\right) \,+\, \sqrt{2 \eta} \, \nu_i^t\;, \label{eq:onlinediscrete-bm-bhid}\\
 \Delta b_j^{\text{vis}} &= \eta \, N\left(x_j^n - \hat{x}_j^n \right) \,+\, \sqrt{2 \eta} \, \nu_j^t\;. \label{eq:onlinediscrete-bm-bvis} 
 \; 
\end{align} 
Note that the parameter prior does not show up in these equations since no priors were used for the biases in our experiments. 
Contrastive divergence approximates the log likelihood derivatives for the weights $w_{ij}$ as $\dd{w_{ij}} \log p_{\mathcal{N}} (\ve x^n | \theta) \approx \, z_i^n x_j^n - \hat{z}_i^n \hat{x}_j^n$. This leads to the synaptic sampling rule
\begin{align}
 \Delta w_{ij} &= \eta \, \left( \dd{w_{ij}} \log p_{\mathcal{S}}(\bw) + N\left(z_i^n x_j^n - \hat{z}_i^n \hat{x}_j^n\right) \right) \,+\, \sqrt{2 \eta} \, \nu_{ij}^t\;. \label{eq:onlinediscrete-bm-w}
\end{align}
In the simulations, we used this rule with $\eta=10^{-4}$ and $N=100$.  Learning started from random initial parameters drawn from a Gaussian distribution with standard deviation $0.25$ and means at $0$ and $-1$ for weights $w_{ij}$ and biases ($b_i^{\text{hid}}$,  $b_j^{\text{vis}}$), respectively.

To compare learning with and without parameter priors, we performed simulations with an uninformative (i.e., uniform) prior on weights (Fig.~\ref{fig:overfitting}C,D), which was implemented by setting $\dd{w_{ij}} \log p_{\mathcal{S}}(\bw)$ to zero.
In simulations with a parameter prior (Fig.~\ref{fig:overfitting}E,F), we used a local prior for each weight in order to obtain local plasticity rules. In other words, the prior $p_\mathcal{S}(\bw)$ was assumed to factorize into priors for individual weights $p_{\mathcal{S}}(\bw) = \prod_{i,j} p_{\mathcal{S}}(w_{ij})$.
For each individual weight prior, we used a bimodal distribution implemented by a mixture of two Gaussians
\begin{align}
  p_{\mathcal{S}}(w_{ij}) \;=\; 0.5\;\Nd\left( w_{ij} \;\middle\vert\; \mu_1, \sigma_1 \right) \;+\; 0.5 \,\Nd\left( w_{ij} \;\middle\vert\; \mu_2, \sigma_2 \right)\;, \label{eqn:rbm_prior} 
\end{align}
with means $\mu_1 = 1.0$, $\mu_2 = 0.0$, and standard deviations $\sigma_1 = \sigma_2 = 0.15$.

\subsection{Details to \textit{Fast adaptation to changing input statistics (Fig.~\ref{fig:structural-plasticity})}}

\subsubsection{Spike-based Winner-Take-All network model}\label{sec:wta}

{\bf Neuron model:} Network neurons were modeled as stochastic spike response neurons with a firing rate that depends exponentially on the membrane voltage \cite{JolivetETAL:06,MensiETAL:11}. 
The membrane potential $u_k(t)$ of neuron $k$ is given by
\begin{equation}
   u_k(t) = \sum_{i} w_{ki} \; x_i(t) \;+\; \beta_k(t)\;, \label{eq:membrane-potential}
\end{equation}
where $x_i(t)$ denotes the (unweighted) input from input neuron $i$, $w_{ki}$ denotes the efficacy of the synapse from input neuron $i$, and $\beta_k(t)$ denotes a homeostatic adaptation current (see below). The input $x_i(t)$ models the influence of additive excitatory postsynaptic potentials (EPSPs) on the membrane potential of the neuron. Let $t_i^{(1)}, t_i^{(2)}, \dots$ denote the spike times of input neuron $i$. Then, $x_i(t)$ is given by
\begin{align}
	x_i(t) = \sum_f \epsilon(t-t_i^{(f)}),
\end{align}
where $\epsilon$ is the response kernel for synaptic input, i.e., the shape of the EPSP, that had a double-exponential form in our simulations:
\begin{equation}
   \epsilon(s) = \Theta(s) \left( e^{-\frac{s}{\tau_f}} - e^{-\frac{s}{\tau_r}} \right)\;,
\label{eq:double-exp-kernel}
\end{equation}
 with the rise-time constant $\tau_r=2$ms, the fall-time constant $\tau_f=20$ms. $\Theta(\cdot)$ denotes the Heaviside step function. 
The instantaneous firing rate $\rho_k(t)$ of network neuron $k$ depends exponentially on the membrane potential and is subject to divisive lateral inhibition $I_{\textnormal{lat}}(t)$ (described below):
\begin{align}
 \rho_k(t) &= \frac{\rho_{\textnormal{net}}}{I_{\textnormal{lat}}(t) } \exp(u_k(t))\;, \label{eq:network-output-rate}
\end{align}
where $\rho_{\textnormal{net}}=100$Hz scales the firing rate of neurons. Such exponential relationship between the membrane potential and the firing rate has been proposed as a good approximation to the firing properties of cortical pyramidal neurons \cite{JolivetETAL:06}.
Spike trains were then drawn from independent Poisson processes with instantaneous rate $\rho_k(t)$ for each neuron. We denote the resulting spike train of the $k^\text{th}$ neuron by $S_k(t)$.
\vspace{0.2cm}

\noindent
{\bf Homeostatic adaptation current:} 
Each output spike caused a slow depressing current, giving rise to the adaptation current $\beta_k(t)$. This implements 
a slow homeostatic mechanism that regulates the output rate of individual neurons (see \cite{HabenschussETAL:12} for details). 
It was implemented as
\begin{align}
\beta_k(t) \,=\, \gamma \sum_f  \kappa(t-t_k^{(f)}),
\end{align}
where $t_k^{(f)}$ denotes the $f$-th spike of neuron $k$ and $\kappa$ is an adaptation kernel that was modeled as a double exponential (Eq.~\eqref{eq:double-exp-kernel}) with time constants $\tau_{r}=12s$ and $\tau_{f} = 30s$. The scaling parameter $\gamma$ was set to $\gamma = -8$.
\\[0.01cm]

\noindent
{\bf Lateral inhibition:} 
Divisive inhibition \cite{Carandini:12} between the $K$ neurons in the WTA network was implemented in an idealized form \cite{NesslerETAL:13}
\begin{align}
  I_{\textnormal{lat}}(t) = \sum_{l=1}^{K} \exp(u_l(t)).
\end{align}
This form of lateral inhibition, that assumes an idealized access to neuronal membrane potentials, has been shown to implement a well-defined generative network model \cite{NesslerETAL:13}, see below.

\subsubsection{Synaptic sampling in spike-based Winner-Take-All networks as stochastic STDP}

It has been shown in \cite{HabenschussETAL:13} that the WTA-network defined above implicitly defines a generative model that is a mixture of Poissonian distributions. In this generative model, inputs $\ve x^n$ are assumed to be generated in dependence on the value of a hidden multinomial random variable $h^n$ that can take on $K$ possible values $1, \dots, K$. Each neuron $k$ in the WTA circuit corresponds to one value $k$ of this hidden variable. In the generative model, for a given value of $h^n=k$, the value of an input $x_i^n$ is then distributed according to a Poisson distribution with a mean that is determined by the synaptic weight $w_{ki}$ from input neuron $i$ to network neuron $k$:
\begin{align}
	p_\mathcal{N}(x_i^n | h^n=k, \ve w) = \Pd(x^n_i | \alpha e^{w_{ki}}), 
\end{align}
with a scaling parameter $\alpha>0$. In other words, the synaptic weight $w_{ki}$ encodes (in log-space) the firing rate of input neuron $i$, given that the hidden cause is $k$. For a given hidden cause, inputs are assumed to be independent, hence one obtains the probability of an input vector for a given hidden cause as  
\begin{align}
	p_\mathcal{N}(\ve x^n | h^n=k, \ve w) = \prod_i \Pd(x^n_i | \alpha e^{w_{ki}}). 
\end{align}
The network implements inference in this generative model, i.e., for a given input $\ve x^n$, the firing rate of network neuron $z_k$ is proportional to the posterior probability $p(h^n=k | \ve x^n, \ve w)$ of the corresponding hidden cause. An online maximum likelihood learning rule for this generative model was derived in \cite{HabenschussETAL:13}. It changes synaptic weights according to
\begin{align}\label{eq:wta_dd_like}
\dd{w_{ki}} \log p_{\mathcal{N}}(\ve x^n \,|\, \ve w)  \approx  S_k(t) \, \left(x_i(t) - \alpha\,e^{w_{ki}}\right)\;,
 \end{align}
where $S_k(t)$ denotes the spike train of the postsynaptic neuron and $x_i(t)$ denotes the weight-normalized value of the sum of EPSPs from presynaptic neuron $i$ at time $t$  (i.e., the summed EPSPs that would arise for weight $w_{ki}=1$;  see Section \ref{sec:wta} for details).
To define the synaptic sampling learning rule completely, we also need to define the parameter prior.
In our experiments, we used a simple Gaussian prior on each parameter $p_\mathcal{S}(\ve \theta) = \prod_{k,i} \Nd(\theta_{ki} | \mu, \sigma^2)$ with $\mu=0.5$ and $\sigma=1$. The derivative of the log-prior is given by    
\begin{align} \label{eq:wta_dd_prior}
\dd{\theta_{ki}} \,\log p_\mathcal{S}(\ve \theta) \;=\; \frac{1}{\sigma^2}\left(\theta_{ki} - \mu \right).
 \end{align}
Inserting Eqs.~\eqref{eq:wta_dd_like} and \eqref{eq:wta_dd_prior} into the general form \eqref{eq:structsamp-general}, we find that the synaptic sampling rule is given by
\begin{align}
 d \theta_{ki} \;=\; b\,\left( \frac{1}{\sigma^2}\left(\theta_{ki} - \mu \right) \;+\;
N w_{ki}\,  S_k(t) \, \left(x_i(t) - \alpha\,e^{w_{ki}}\right) \right) dt  \; +\sqrt{2b} \, d \wiener_{ki} \;,
\label{eqn:sup-sem-rule}
\end{align}
which corresponds to rule \eqref{eq:structsamp-spiking} with double indices $ki$ replaced by single parameter indexing $i$ to simplify notation.

\subsubsection{Simulation details for spiking network simulations (Figs.~\ref{fig:prior-bars}, \ref{fig:structural-plasticity}, and \ref{fig:self-repair})}

Computer simulations of spiking neural networks were based on adapted event-based simulation software from \cite{KappelETAL:14}.
In all spiking neural network simulations, synaptic weights were updated according to the rule \eqref{eq:structsamp-spiking} with parameters $N=100$, $\alpha=e^{-2}$, and $b = 10^{-4}$, except for panel \ref{fig:prior-bars}F where $b=10^{-6}$ was used as a control. In the simulations, we directly implemented the time-continuous evolution of the network parameters in an event-based update scheme. 
Before learning, initial synaptic parameters were independently drawn from the prior distribution $p_\mathcal{S}(\ve \theta)$.

For the mapping \eqref{eq:thetamap} from synaptic parameters $\theta_{ki}$ to synaptic efficacies $w_{ki}$, we used as offset $\theta_0=3$.
 This results in synaptic weights that shrink to small values ($< 0.05$) when synaptic parameters are below zero. In the simulation, we clipped the synaptic weights to zero for negative synaptic parameters $\theta$ to account for retracted synapses. More precisely, the actual weights $\hat{w}_{ki}$ used for the computation of the membrane potential \eqref{eq:membrane-potential} were given by $\hat{w}_{ki} \,=\, \max \left\{ 0,  w_{ki} \,-\, \exp(-\theta_0) \right\}\;$.
 To avoid numerical problems, we clipped the synaptic parameters at $-5$ and the maximum amplitude of instantaneous parameter changes to $5 b$.

\subsubsection{Network inputs for Figs.~\ref{fig:prior-bars} and \ref{fig:structural-plasticity}}
Handwritten digit images were taken from the MNIST dataset \cite{LeCunETAL:98}. 
Each pixel was represented by a single afferent neuron. Gray scale values where scaled to 0-50Hz Poisson input rate and 1Hz input noise was added on top. These Poisson rates were kept fixed for each example input digit for a duration of $200$ms. After each digit presentation, a $50$ms window of $1$Hz Poisson noise on all input channels was presented before the next digit was shown. 

\subsection{Details to \textit{Inherent compensation capabilities of networks with synaptic sampling (Fig.~\ref{fig:self-repair})}}
\label{sec:meth-self-repair}

Here we provide details to the network model and spiking inputs for the recurrent WTA circuits. Additional details to the data analysis and performance evaluation are provided in \nameref{sec:supplement}.

\subsubsection{Network model}

In Fig.~\ref{fig:self-repair} two recurrently connected ensembles, each consisting of four WTA circuits, were used. The parameters of neuron and synapse dynamics were as described in Section \ref{sec:wta}. All synapses, lateral and feedforward, were subject to the same learning rule \eqref{eq:structsamp-spiking}. Lateral connections within and between the WTA Circuit neurons were unconstrained (allowing potentially all-to-all connectivity). Connections from input neurons were constraint as shown in Fig.~\ref{fig:self-repair}. The lateral synapses were treated in the same way as synapses from input neurons but had a synaptic delay of $5$ms.

\subsubsection{Network inputs}
The spoken digit presentations were given by reconstructed cochleagrams of speech samples of isolated spoken digits from the TI~46 dataset  (also used in \cite{KlampflMaass:13,HopfieldBrody:01}). Each of the 77 channels of the cochleagrams was represented by 10 afferent neurons, giving a total of 770. Cochleagrams were normalized between 0 and 80Hz and used to draw individual Poisson spike trains for each afferent neuron.  In addition 1Hz Poisson noise was added on top. We used 10 different utterances of digits \textit{1} and \textit{2} of a single speaker. We selected 7 utterances for training and 3 for testing. For training, one randomly selected utterance was presented together with a randomly chosen instance of the corresponding handwritten digit. The spike patterns for the written digits were generated as in Section \ref{sec:adaptation} and had the same duration as the spoken digits. Each digit presentation was padded with $25$ms, $1$Hz Poisson noise before and after the digit pattern.

For test trials in which only the auditory stimulus was presented, the activity of the visual input neurons was set to $1$Hz throughout the whole pattern presentation. The learning rate $b$ was set to zero during these trials. The PETH plots were computed over 100 trial responses of the network to the same stimulus class (e.g. presentation of digit \textit{1}). Spike patterns for input stimuli were randomly drawn in each trial for the given rates. Spike trains were then filtered with a Gaussian filter with $\sigma = 50$ms and summed in a time-discrete matrix with $10$ms bin length. Maximum firing times were assigned to the time bin with the highest PETH amplitude for each neuron.

\section{Supplement}
\label{sec:supplement}
The supplement is available on the author's web page \url{http://www.igi.tugraz.at/kappel/}.

\section*{Acknowledgments}
We would like to thank Seth Grant, Christopher Harvey, Jason MacLean and Simon Rumpel for helpful comments.

\bibliography{refs,/home/bibs/igi,/home/bibs/nnrefs}
\clearpage

\end{document}

%% file: new_commands.tex
\newcommand{\ve}[1]{\boldsymbol{#1}}

\newtheorem{thm}{Theorem}

\definecolor{gray}{gray}{0.2}



\newcommand{\wiener}{\mathcal{W}}